\documentclass[10pt,twocolumn,letterpaper]{article}

\usepackage{multirow}
\usepackage{bbm}
\usepackage[pagenumbers]{cvpr} % To force page numbers, e.g. for an arXiv version

\usepackage{multirow}
\usepackage{colortbl} 
\usepackage{booktabs} 
\usepackage{makecell}
\definecolor{cvprblue}{rgb}{0.21,0.49,0.74}
\usepackage[pagebackref,breaklinks,colorlinks,allcolors=cvprblue]{hyperref}

\def\confName{CVPR}
\def\confYear{2027}
\definecolor{vicalight}{RGB}{244, 247, 250}
\definecolor{vicaplus}{RGB}{232, 240, 249}

\title{VICA: Video Contextual Agreement for AIGC Video Detection and Localization}

\author{\textbf{Runheng Lin}$^{2,}$\thanks{Equal contribution}\quad\,
\textbf{Jialiang Shen}$^{3,}$\footnotemark[1]\quad
\textbf{Jiyang Zheng}$^{4}$\quad
\textbf{Ruiqi Liu}$^{5}$\vspace{0.25em}\\
\textbf{Zonglin He}$^{6}$\quad
\textbf{Yu Yao}$^{3}$\quad
\textbf{Huajie Chen}$^{7}$\quad
\textbf{Chris Russell}$^{1}$\quad
\textbf{Runqi Lin}$^{1,}$\thanks{Corresponding author: runqi.lin@oii.ox.ac.uk}
\vspace{0.5em}
\\
$^{1}$University of Oxford \quad
$^{2}$Xiamen University \quad 
$^{3}$The University of Sydney\vspace{0.25em}\\
$^{4}$Microsoft \quad 
$^{5}$Institute of Automation, Chinese Academy of Sciences \vspace{0.25em}\\
$^{6}$ Baidu \quad
$^{7}$City University of Macau 
}

\begin{document}

\maketitle
\begingroup
\renewcommand{\thefootnote}{}
\footnotetext[0]{A Preprint.}
\endgroup
\begin{abstract}
The rapid advancement of AI-generated content (AIGC) video generation and editing technologies has raised growing societal concerns about the authenticity and reliability of digital videos.
Existing video detection methods primarily exploit temporal inconsistencies through holistic frame-level representations, whereas manipulation localization methods focus on artifacts within individual spatial regions.
However, holistic representations may obscure localized artifacts through global feature aggregation, while region-centric approaches may overlook disagreements between individual regions and their surrounding spatiotemporal context.
To address these limitations, we propose \textbf{VI}deo \textbf{C}ontextual \textbf{A}greement (VICA), a unified framework that masks individual video regions and leverages a frozen V-JEPA model to predict their representations from the surrounding spatiotemporal context.
Complementary masking ensures complete regional coverage, while context-conditioned predictions establish references for assessing regional agreement with the broader video context.
By comparing predicted and observed representations, VICA derives spatial residuals that characterize intra-frame discrepancies and bidirectional temporal residuals that capture inconsistencies across preceding and subsequent frames.
These residuals characterize contextual disagreements, providing a shared forensic representation for both tasks and shifting the focus from identifying isolated manipulation artifacts to assessing regional agreement with expectations inferred from the broader video context.
Extensive experiments demonstrate that VICA achieves state-of-the-art performance on both tasks across multiple benchmarks, with strong generalization to unseen video generators and editing methods.

% The rapid advancement of AI video generation and editing technologies has raised growing societal concerns about video authenticity, particularly the potential for manipulated content to mislead the public.
% Existing detection methods primarily leverage holistic frame representations to capture intra-frame visual artifacts or inconsistencies across adjacent frames.
% However, such holistic representations may dilute local artifacts with irrelevant visual information and overlook disagreements between individual regions and the broader video context.
% To this end, we propose \textbf{VI}deo \textbf{C}ontextual \textbf{A}greement (VICA), a unified framework for AI-generated video detection and AI-edited region localization that evaluates contextual agreement at the region level.
% Specifically, VICA employs complementary masks and a frozen V-JEPA model to predict each masked region from its visible spatiotemporal context.
% Such region-by-region prediction preserves local evidence that holistic representations tend to dilute, and each prediction is informed by both the current frame and the content changes across the clip.
% Comparing the predictions with the observed features, VICA derives spatial residuals that reveal discrepancies within the current frame and bidirectional temporal residuals that reveal deviations from past and future frames.
% Extensive experiments demonstrate that VICA achieves state-of-the-art performance on both tasks across multiple benchmarks, while exhibiting strong generalization to unseen video generators.

\end{abstract}
    
\section{Introduction}
\label{sec:intro}

\begin{figure*}[t]
    \centering
    \includegraphics[width=1.\textwidth]{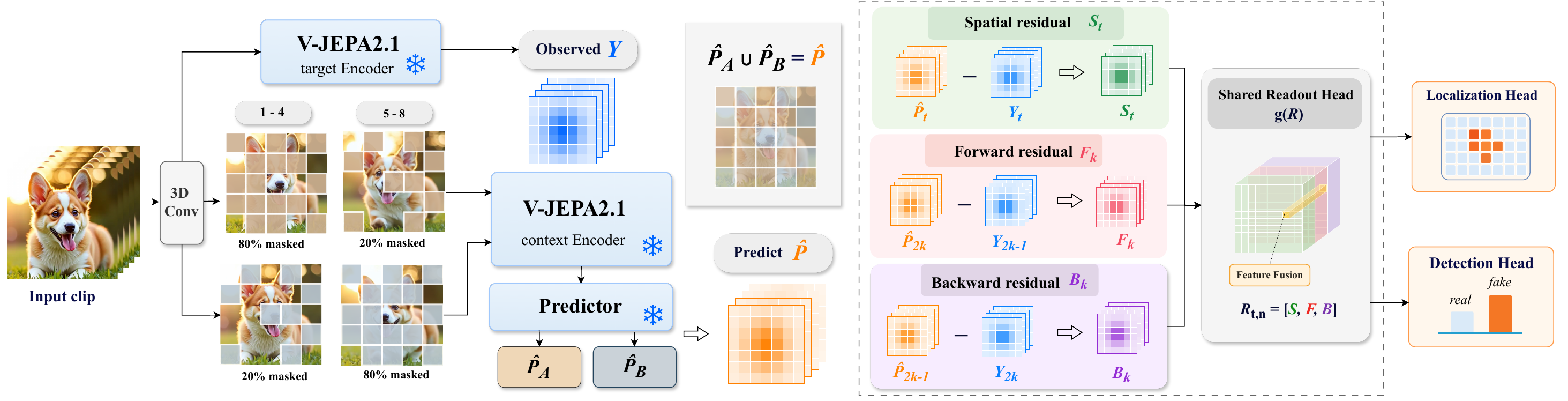}
    \vspace{-0.5em}
    \caption{\textbf{Overview of VICA.} Given an input video clip, the frozen V-JEPA~2.1 target encoder extracts observed features $Y$, while the frozen context encoder and predictor generate context-conditioned predictions from two complementary masked views. The resulting predictions are combined into $\widehat{P}$, providing a predicted representation for every spatiotemporal token. Spatial residuals $S$ are computed between predicted and observed features at the same time step, whereas bidirectional temporal residuals $F$ and $B$ capture discrepancies across paired time steps. Finally, a shared readout integrates $[S,F,B]$ to support both video-level detection and region-level localization.}
    \vspace{-1em}
    \label{fig:core-framework}
\end{figure*}

Recent advances in AI-driven video generation~\citep{wu2025hunyuanvideo,seedance2026seedance,ICLR2026_41b93c59,videoworldsimulators2024,hong2022cogvideo,zheng2024open,wan2025wan,fan2024freemotion} and editing~\citep{geyer2024tokenflow,liu2024video,ye2025unic,jiang2025vace,yang2026videocof} have enabled the synthesis of photorealistic videos and the seamless modification of selected video regions. These advances have expanded the possibilities for video content creation and post-production, enabling realistic visual effects and more flexible content manipulation~\citep{zhang2025generative, azzarelli2025intelligent}. However, the increasing realism of synthetic and edited videos has made it increasingly difficult to distinguish authentic content from AI-generated or manipulated content. This poses significant challenges to video authenticity verification, particularly in high-stakes scenarios such as legal proceedings, where manipulated footage may compromise the integrity of visual evidence~\citep{mcconvey2026benchmark}. Addressing these challenges requires forensic methods capable of both detecting AI-generated videos and localizing manipulated regions.

Existing approaches to AI-generated video detection primarily exploit temporal inconsistencies in holistic visual representations.
For instance, D3~\cite{zheng2025d3} examines second-order differences between frame representations, while ReStraV~\cite{interno2026ai} characterizes the curvature of their temporal trajectories.
TRACE~\cite{cao2026trace} instead analyzes the velocity responses of a pretrained video generator across noise levels and neighboring frames.
Despite their effectiveness, these approaches may overlook localized manipulation cues, as holistic representations can obscure subtle regional discrepancies.
V-PVP~\cite{cui2026rethinking} further demonstrates the importance of preserving fine-grained forensic information during feature aggregation.
Meanwhile, existing video manipulation localization methods~\cite{lou2025video,montibeller2026ainpaint} primarily identify artifacts within individual regions, without explicitly assessing their agreement with the surrounding spatiotemporal context.
Consequently, visually plausible manipulations that exhibit contextual inconsistencies may remain undetected.
These limitations highlight the need for a unified forensic approach that preserves localized evidence while explicitly assessing contextual agreement at the regional level.

\begin{figure*}[t]
    \centering
    \includegraphics[width=1.\textwidth]{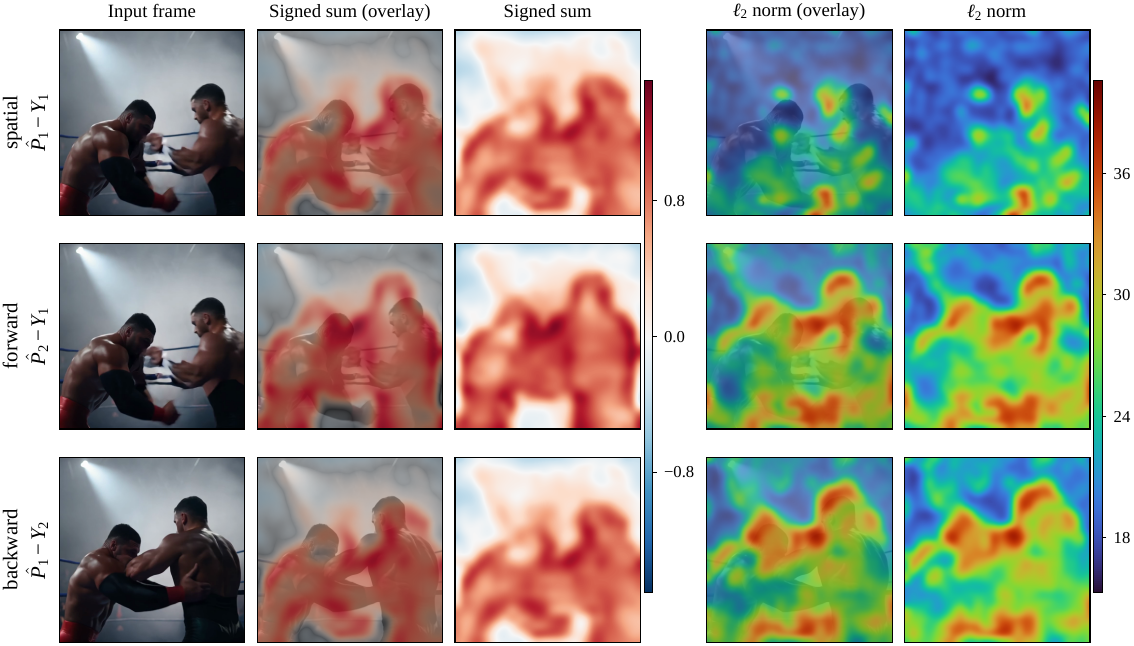}
    \caption{The video is generated by Vidu and sourced from AIGVDBench, which does not provide localization annotations. $\widehat{P}_t$ and $Y_t$ denote the predicted and observed regional features at temporal step $t$, respectively. The three rows show the spatial residual $\widehat{P}_1-Y_1$ and the bidirectional temporal residuals $\widehat{P}_2-Y_1$ and $\widehat{P}_1-Y_2$. The columns display the input frame, channel-wise signed residual sums, and $\ell_2$ norms, with and without overlays on the input frame. The right boxer’s glove appears malformed, while the spatial residual $\widehat{P}_1-Y_1$ produces a localized response around the fist. In comparison, the bidirectional temporal residuals $\widehat{P}_2-Y_1$ and $\widehat{P}_1-Y_2$ exhibit stronger and more extensive $\ell_2$ responses along the moving arms and gloves, highlighting motion-dependent discrepancies that are less apparent in the spatial residual. This visualization illustrates how spatial and bidirectional temporal residuals reveal complementary patterns of contextual disagreement in AI-generated videos.}
    \label{fig:residual-0001}
    \vspace{-1em}
\end{figure*}

To this end, we propose \textbf{VI}deo \textbf{C}ontextual \textbf{A}greement (VICA), a unified framework for AIGC video detection and AI-edited region localization. Built upon a frozen V-JEPA 2.1~\cite{mur2026v}, VICA establishes context-conditioned references for individual video regions by predicting their representations from the surrounding spatiotemporal context. To obtain predictions for every region, we introduce a complementary masking strategy that constructs two masked views with asymmetric masking ratios across the temporal halves of the video. This design ensures complete regional coverage while encouraging predictions to leverage temporally distant context. VICA then assesses contextual agreement by comparing predicted and observed representations, deriving spatial residuals that characterize discrepancies within individual temporal steps and bidirectional temporal residuals that capture inconsistencies between preceding and subsequent temporal steps. As illustrated in \cref{fig:residual-0001}, these residuals reveal complementary spatial and temporal patterns of contextual disagreement in AI-generated videos. Finally, a shared readout processes the residual representations and feeds them into separate video-level detection and region-level localization heads. By relying exclusively on context-conditioned residuals, VICA provides a unified forensic representation for both tasks without requiring separate task-specific feature extraction pipelines.

Our main contributions are summarised as follows:
\begin{itemize}[leftmargin=11pt, topsep=2pt, itemsep=2pt]
\item We introduce \emph{contextual agreement} as a forensic cue for AI-generated video detection and AI-edited region localization, shifting the focus from isolated artifacts to assessing regional consistency with expectations inferred from the broader spatiotemporal context.

\item We propose VICA, a unified framework built upon a frozen V-JEPA~2.1 model that employs complementary masking to generate context-conditioned predictions for every video region and uses spatial and bidirectional temporal residuals to characterize contextual disagreements, providing a shared forensic representation for both tasks.

\item Extensive experiments demonstrate that VICA achieves strong detection and localization performance across multiple benchmarks, with substantial improvements over existing methods and robust generalization to unseen video generators and editing methods.

\end{itemize}

\section{Related Work}
\label{sec:related-work}

\subsection{AIGC Image Detectors}

AI-generated image detectors can be adapted for video detection by aggregating predictions across individual frames. UniFD~\cite{ojha2023towards} trains a linear classifier on frozen CLIP features to improve generalization across image generators. NPR~\cite{tan2024rethinking} exploits local upsampling artifacts through neighboring pixel relationships, while DDA~\cite{chen2025dual} aligns real and synthetic images in the spatial and frequency domains to mitigate detection biases. Effort~\cite{yan2024effort} employs singular value decomposition to preserve pretrained knowledge in a frozen principal subspace while learning forgery-specific representations in its orthogonal complement. MIRROR~\cite{liu2026mirror} reconstructs image representations from a memory of real-image prototypes to identify synthetic content through reconstruction residuals. Although these methods provide useful image-level forensic cues, their frame-wise application to videos does not explicitly model temporal dependencies, limiting their ability to capture inconsistencies across frames.

\subsection{AIGC Video Detection}

To capture temporal inconsistencies, existing AIGC video detection methods exploit temporal dynamics in video representations. D3~\cite{zheng2025d3} examines second-order differences in frame representations, while STALL~\cite{ben2026training} models the distributions of real-video frame embeddings and their temporal transitions to identify unlikely sequences. ReStraV~\cite{interno2026ai} characterizes changes in the magnitude and direction of inter-frame feature differences, whereas TRACE~\cite{cao2026trace} analyzes temporal variations in the velocity responses of a pretrained video generator. Beyond these approaches, ReConFuse~\cite{chen2026reconfuse} integrates reconstruction errors from a pretrained video autoencoder with semantic features and temporal modeling. While these methods provide useful forensic cues, their reliance on frame-level representations or aggregated signals may obscure localized inconsistencies. To preserve fine-grained evidence, V-PVP~\cite{cui2026rethinking} introduces velocity-gated patch aggregation, demonstrating the importance of retaining patch-level dynamics during video-level detection. Nevertheless, preserving regional information alone does not explicitly establish its agreement with the surrounding spatiotemporal context.

\subsection{AI-Edited Video Localization}

AI-edited video localization aims to identify manipulated regions within otherwise authentic videos. ViLocal~\cite{lou2025video} learns spatiotemporal forensic representations through pixel-level contrastive learning to distinguish inpainted regions from authentic content. AINPAINT~\cite{montibeller2026ainpaint} introduces a practical video-inpainting dataset and a dual-branch architecture for manipulation localization. While these methods extract forensic cues from local visual representations, they do not explicitly assess regional agreement with the surrounding spatiotemporal context. More recently, RelayFormer~\cite{huang2026relayformer} employs global–local relay attention to facilitate contextual information exchange across spatial and temporal regions. However, its reliance on a limited set of relay tokens for global information aggregation may introduce an information bottleneck, potentially obscuring fine-grained cross-region dependencies. VICA instead explicitly evaluates contextual agreement at each spatiotemporal region through prediction–observation residuals, preserving localized forensic evidence while incorporating information inferred from the broader video context.
\section{Method}
\label{sec:method}

In this section, we present VICA, a unified framework for AIGC video detection and AI-edited region localization that evaluates \emph{contextual agreement} by comparing regional observations with expectations inferred from the surrounding spatiotemporal context. As illustrated in \cref{fig:core-framework}, VICA first extracts observed features from the input clip and generates context-conditioned regional predictions using complementary masked views (\cref{sec:method-features}). It then computes spatial and bidirectional temporal residuals between predicted and observed features, combining them into region-level residual representations (\cref{sec:method-residuals}). Finally, a shared readout processes these residual representations to support video-level detection and region-level localization (\cref{sec:method-heads}).

\subsection{Features and Complementary Masks}
\label{sec:method-features}

\noindent\textbf{Observed Features.}\hspace{2mm}Following V-JEPA 2.1~\cite{mur2026v}, we denote the frozen context encoder, target encoder, and predictor as $E_{\theta}$, $E_{\bar{\theta}}$, and $P_{\phi}$, respectively. Given a 16-frame video clip $x$, 3D tubelet tokenization yields $T=8$ temporal steps, each spanning two consecutive frames and containing $N$ spatial tokens. The first and second halves of the clip correspond to temporal steps $1$–$4$ and $5$–$8$, respectively. Each token $(t,n)$, where
$t\in\{1,\ldots,T\}$ and
$n\in\{1,\ldots,N\}$,
represents a local spatiotemporal region. We extract observed features at four feature levels by passing the unmasked clip through the target encoder:
\begin{equation}
Y^{l} = E_{\bar{\theta}}^{l}(x)
\in \mathbb{R}^{T\times N\times D},
\quad l=1,\ldots,4.
\label{eq:observed}
\end{equation}
Here, $E_{\bar{\theta}}^{l}$ denotes the target encoder output at feature level $l$, $D$ is the feature dimension, and $Y_t^{l}\in\mathbb{R}^{N\times D}$ and $Y_{t,n}^{l}\in\mathbb{R}^{D}$ denote the observed features at temporal step $t$ and token $(t,n)$, respectively.

\noindent\textbf{Complementary Masks.}\hspace{2mm}To obtain context-conditioned predictions for every region, we construct two complementary masked views $\mathrm{A}$ and $\mathrm{B}$ with binary masks $M_{\mathrm{A}},M_{\mathrm{B}}\in{0,1}^{T\times N}$ satisfying $M_{\mathrm{B}}=\mathbbm{1}-M_{\mathrm{A}}$, where $\mathbbm{1}$ denotes the all-ones matrix. A mask entry $M_{v,t,n}=1$ indicates that token $(t,n)$ is hidden in view $v\in{\mathrm{A},\mathrm{B}}$. Consequently, every token is masked in exactly one view, ensuring that a context-conditioned prediction is obtained for every region.

We then adopt an asymmetric masking schedule across the two temporal halves of the clip. Let $\rho_v(t)=\frac{1}{N}\sum_{n=1}^{N}M_{v,t,n}$ denote the masking ratio at temporal step $t$ in view $v$. The masking ratios are defined as
\begin{equation}
\begin{aligned}
\rho_{\mathrm{A}}(t)
&=
\begin{cases}
0.2, & t \le T/2,\\
0.8, & t > T/2,
\end{cases}\\
\rho_{\mathrm{B}}(t)
&= 1-\rho_{\mathrm{A}}(t).
\end{aligned}
\label{eq:mask-ratio}
\end{equation}
This asymmetric schedule encourages predictions to leverage temporally distant context while preserving visible spatial information at every temporal step. Specifically, the higher masking ratio in one temporal half makes the opposite half a richer source of visible context. Furthermore, the same randomly sampled spatial mask is maintained across all temporal steps within each half, preventing the predictor from accessing masked spatial locations at nearby steps and encouraging predictions based on visible spatial neighbors and temporally distant observations.

\noindent\textbf{Regional Feature Prediction.}\hspace{2mm}For each masked view $v\in\{\mathrm{A},\mathrm{B}\}$, the context encoder processes only the visible tokens $x_{\neg M_v}$, where $M_{v,t,n}=0$.
The predictor then combines the encoded context with learnable mask tokens $\Delta_{M_v}$ at the masked positions to predict their corresponding regional features:
\begin{equation}
\widehat{P}_{v}^{l}
=
P_{\phi}^{l}\big(
E_{\theta}(x_{\neg M_v}),
\Delta_{M_v}
\big).
\label{eq:prediction}
\end{equation}
Here, $P_{\phi}^{l}$ denotes the predictor mapping to the output at the $l$-th selected feature level, corresponding to \cref{eq:observed}.
For notational convenience, $\widehat{P}_{v}^{l}\in\mathbb{R}^{T\times N\times D}$ is represented on the full token grid, with entries at visible positions set to zero.
Since the two masks are complementary, their predictions can be combined to obtain a complete set of regional predictions:
\begin{equation}
\widehat{P}^{l}
=
M_{\mathrm{A}}\odot\widehat{P}_{\mathrm{A}}^{l}
+
M_{\mathrm{B}}\odot\widehat{P}_{\mathrm{B}}^{l}.
\label{eq:assembly}
\end{equation}
Here, $\widehat{P}^{l}\in\mathbb{R}^{T\times N\times D}$, and $\odot$ denotes element-wise multiplication with broadcasting along the feature dimension.
Each predicted feature $\widehat{P}_{t,n}^{l}$ is inferred without observing its corresponding input token, providing a context-conditioned reference for comparison with the observed feature $Y_{t,n}^{l}$.

\subsection{Spatial and Bidirectional Temporal Residuals}
\label{sec:method-residuals}

The discrepancy between predicted features $\widehat{P}$ and observed features $Y$ characterizes contextual disagreement at the regional level. However, prediction errors can also arise in authentic videos, making residual magnitude alone insufficient to reliably identify manipulated content. We therefore retain the full residual vectors as forensic representations rather than directly thresholding their magnitudes. The following residuals are computed independently at each feature level, with the superscript $l$ omitted for clarity.

\noindent\textbf{Spatial Residual.}\hspace{2mm}We first compute the spatial residual between predicted and observed features at each spatiotemporal token:
\begin{equation}
S_{t,n} = \widehat{P}_{t,n} - Y_{t,n}.
\label{eq:spatial-residual}
\end{equation}
Computing these discrepancies before spatial aggregation preserves region-specific information that might otherwise be obscured by unrelated visual content.

\noindent\textbf{Bidirectional Temporal Residuals.}\hspace{2mm}
Spatial residuals characterize prediction discrepancies at individual temporal steps but do not explicitly capture temporal variations across steps. To incorporate local temporal dynamics, we group the $T$ temporal steps into consecutive pairs $(2k-1,2k)$, where $k=1,\ldots,T/2$. For each pair, we compute bidirectional temporal residuals by comparing the context-conditioned prediction at one step with the observation at the other:
\begin{equation}
\begin{aligned}
F_{k,n} &= \widehat{P}_{2k,n} - Y_{2k-1,n},\\
B_{k,n} &= \widehat{P}_{2k-1,n} - Y_{2k,n}.
\end{aligned}
\label{eq:temporal-residual}
\end{equation}
Unlike spatial residuals, these cross-step comparisons incorporate observed temporal changes into contextual prediction discrepancies, providing complementary cues for identifying temporal inconsistencies.

% Since all four steps in each half share the same spatial mask, both steps of a pair are hidden in the same view and predicted from the exact same visible tokens, providing a consistent contextual reference across the pair.

% Using both temporal directions is essential to disentangle two distinct failure modes.
% Let $\delta_k = Y_{2k} - Y_{2k-1}$ be the observed temporal change and $\hat{\delta}_k = \widehat{P}_{2k} - \widehat{P}_{2k-1}$ be the context-implied change.
% The two spatial residuals of a pair only expose their difference, $S_{2k} - S_{2k-1} = \hat{\delta}_k - \delta_k$, confounding an unexpected observed change with an expected change that the observation lacks.
% Combining $S_t$ with $F_k$ and $B_k$ separates the two:
% \begin{equation}
%     \begin{aligned}
%         \hat{\delta}_k &= F_k - S_{2k-1} = S_{2k} - B_k, \\
%         \delta_k &= F_k - S_{2k} = S_{2k-1} - B_k.
%     \end{aligned}
%     \label{eq:temporal-identity}
% \end{equation}
% \Cref{eq:temporal-identity} also explains why both directions are needed.
% Because each token holds only its own step's spatial residual, a single temporal direction recovers only one of $\delta_k$ and $\hat{\delta}_k$ at a given step, whereas using both $F_k$ and $B_k$ gives every token access to both changes regardless of its position within the pair (ablated in \cref{tab:residual}).

\subsection{Residual-Only Readout and Task Heads}
\label{sec:method-heads}

\noindent\textbf{Shared Readout.}\hspace{2mm}For each token $(t,n)$, we concatenate its spatial residual
$S_{t,n}$ with the bidirectional temporal residuals
$(F_{k,n},B_{k,n})$ from the corresponding temporal pair
$k=\lceil t/2\rceil$, across all four feature levels:
\begin{equation}
R_{t,n}
=
\big[S^{l}_{t,n}, F^{l}_{k,n}, B^{l}_{k,n}\big]_{l=1}^{4}
\in \mathbb{R}^{12D},
\label{eq:residual}
\end{equation}
where $[\cdot]_{l=1}^{4}$ denotes channel-wise concatenation
across all three residual types and four feature levels.
The resulting residual representation
$R\in\mathbb{R}^{T\times N\times 12D}$ preserves the
original spatiotemporal token structure without spatial or temporal
aggregation.

A shared readout $g$ maps each $12D$-dimensional residual
vector to a 256-dimensional feature space, yielding
$Z=g(R)\in\mathbb{R}^{T\times N\times 256}$, where
$z_{t,n}\in\mathbb{R}^{256}$ denotes the corresponding representation
of token $(t,n)$.
Both task heads operate exclusively on these residual-derived
features.
The localization head applies a linear classifier to each
token representation $z_{t,n}$, while the detection head
aggregates $Z$ using spatial mean and standard deviation,
followed by temporal soft-maximum pooling and a linear
classifier.
The resulting probabilities $p_{t,n}$ and $p$ indicate
whether the respective tubelet $(t,n)$ is manipulated and whether the
input clip is AI-generated, respectively.

\noindent\textbf{Task Training.}\hspace{2mm}We optimize the shared readout $g$ and two task-specific heads while keeping the V-JEPA~2.1 backbone frozen throughout training.
We jointly train these components using a localization batch
$\mathcal{B}_{\mathrm{loc}}$ with tubelet-level annotations
$a_{t,n}\in\{0,1\}$ and a detection batch
$\mathcal{B}_{\mathrm{vid}}$ with video-level labels
$c\in\{0,1\}$.
A tubelet is labeled as manipulated ($a_{t,n}=1$)
if it contains at least one edited pixel.
The training objective combines binary cross-entropy (BCE)
losses for both tasks:
\begin{equation}
\begin{aligned}
\mathcal{L}_{\mathrm{loc}}
&= \frac{1}{|\mathcal{B}_{\mathrm{loc}}|\,TN}
\sum_{j\in\mathcal{B}_{\mathrm{loc}}}
\sum_{t,n}
\operatorname{BCE}\big(p^{(j)}_{t,n},a^{(j)}_{t,n}\big),\\
\mathcal{L}_{\mathrm{vid}}
&= \frac{1}{|\mathcal{B}_{\mathrm{vid}}|}
\sum_{j\in\mathcal{B}_{\mathrm{vid}}}
\operatorname{BCE}\big(p^{(j)},c^{(j)}\big),\\
\mathcal{L}
&= \mathcal{L}_{\mathrm{loc}}+\mathcal{L}_{\mathrm{vid}}.
\end{aligned}
\label{eq:loss}
\end{equation}
Here, the superscript $(j)$ indexes individual samples,
$\mathcal{L}_{\mathrm{loc}}$ is averaged over all
spatiotemporal tokens in the localization batch, and
$\mathcal{L}_{\mathrm{vid}}$ is averaged over videos in
the detection batch.
For single-task training, only the corresponding data
stream and task head are used.
\section{Experiments}
\label{sec:experiments}

% 导言区需要 \usepackage{multirow}
\begin{table*}[t]
\centering
\caption{AIGC video detection results in AUROC and AP (\%).
Following the AIGVDBench protocol, trainable detectors are
trained exclusively on real and Open-Sora videos from
AIGVDBench and evaluated across all listed benchmarks.
Metrics are computed separately for each generator against
the corresponding real videos and averaged across generators.
AIGVDBench includes 20 open-source generators (including
Open-Sora) and 11 closed-source generators.
VICA-Det uses only AIGVDBench detection supervision,
whereas the unified VICA additionally incorporates
AINPAINT localization supervision.
\textbf{Bold} and \underline{underlined} denote the best
and second-best results.}
\vspace{-0.5em}
\label{tab:core-main}

\setlength{\tabcolsep}{4.5pt}
\setlength{\extrarowheight}{1.5pt}
\renewcommand{\arraystretch}{1.12}
\setlength{\aboverulesep}{0pt}
\setlength{\belowrulesep}{0pt}
\resizebox{\textwidth}{!}{%
\begin{tabular}{l cccc cc cc cc cc}
\toprule
\multirow{3}{*}{Method} & \multicolumn{4}{c}{AIGVDBench} & \multicolumn{2}{c}{\multirow{2}{*}{GenVidBench}} & \multicolumn{2}{c}{\multirow{2}{*}{VideoFeedback}} & \multicolumn{2}{c}{\multirow{2}{*}{Magic Videos}} & \multicolumn{2}{c}{\multirow{2}{*}{GVF}}\\
\cmidrule(lr){2-5}
 & \multicolumn{2}{c}{Open-source} & \multicolumn{2}{c}{Closed-source} & & & & & & & & \\
\cmidrule(lr){2-3}\cmidrule(lr){4-5}\cmidrule(lr){6-7}\cmidrule(lr){8-9}\cmidrule(lr){10-11}\cmidrule(lr){12-13}
 & AUROC & AP & AUROC & AP & AUROC & AP & AUROC & AP & AUROC & AP & AUROC & AP\\
\midrule
D3~\cite{zheng2025d3} & 64.88 & 64.66 & 71.53 & 72.92 & 81.87 & 74.75 & 65.06 & 69.13 & 67.69 & 69.39 & \underline{78.51} & \underline{75.99}\\
DeMamba~\cite{chen2026demamba} & 72.43 & 73.66 & 62.86 & 65.38 & 52.63 & 55.44 & 55.00 & 68.75 & 43.21 & 44.35 & 26.07 & 39.86\\
DeCoF~\cite{ma2025detecting} & 90.10 & 89.73 & 75.82 & 75.50 & 49.32 & 52.03 & 51.79 & 62.85 & 44.84 & 48.79 & 21.39 & 36.69\\
ReStraV~\cite{interno2026ai} & 57.02 & 59.54 & 40.75 & 42.81 & 91.50 & \underline{92.31} & 48.64 & 58.98 & 49.84 & 51.14 & 52.07 & 50.74\\
SPLIT~\cite{hyun2026split} & 41.33 & 48.32 & 43.64 & 48.78 & 68.78 & 64.01 & 52.38 & 56.15 & 41.38 & 45.10 & 38.22 & 45.63\\
WaveRep~\cite{corvi2025seeing} & 76.38 & 77.74 & 83.72 & 83.84 & 49.55 & 51.92 & 35.30 & 46.83 & 55.98 & 52.14 & 24.09 & 37.05\\
STALL~\cite{ben2026training} & 44.60 & 47.17 & 77.21 & 77.34 & 42.12 & 45.13 & 16.77 & 33.91 & 71.32 & 67.07 & 62.19 & 63.01\\
V-PVP~\cite{cui2026rethinking} & 88.34 & 87.93 & \underline{88.28} & 86.92 & \underline{92.04} & 88.73 & 68.76 & 68.84 & 63.65 & 61.67 & 76.67 & 73.38\\
\midrule
NPR~\cite{tan2024rethinking} & 56.76 & 57.64 & 67.74 & 67.32 & 48.19 & 55.12 & 53.10 & 52.65 & 74.34 & 72.69 & 68.21 & 68.64 \\
DDA~\cite{chen2025dual} & 89.47 & 87.77 & 87.42 & \underline{88.23} & 72.62 & 79.47 & 72.96 & 75.07 & 81.33 & 77.66 & 58.64 & 65.37\\
UniFD~\cite{ojha2023towards} & 56.50 & 56.76 & 53.79 & 53.77 & 62.34 & 67.25 & 67.80 & 71.92 & 51.19 & 51.08 & 63.46 & 65.95\\
Effort~\cite{yan2024effort} & 72.03 & 72.01 & 64.60 & 64.49 & 48.70 & 57.16 & 50.00 & 53.66 & 52.22 & 54.05 & 24.96 & 43.58\\
\midrule
\rowcolor{vicalight}
VICA-Det & \textbf{94.74} & \textbf{94.15} & 85.89 & 85.20 & 87.29 & 85.79 & \underline{84.71} & \underline{88.57} & \textbf{83.83} & \underline{80.80} & 64.55 & 69.81\\
\rowcolor{vicaplus}
VICA-Unified & \underline{93.52} & \underline{93.49} & \textbf{90.13} & \textbf{90.29} & \textbf{92.71} & \textbf{92.50} & \textbf{90.69} & \textbf{92.51} & \underline{83.56} & \textbf{82.49} & \textbf{87.01} & \textbf{80.80}\\
\bottomrule
\end{tabular}%
}
\vspace{-0.5em}

\end{table*}

\noindent\textbf{Overview.}\hspace{2mm}We evaluate VICA on AIGC video detection and AI-edited region
localization across multiple benchmarks, emphasizing
generalization to unseen video generators and editing methods.
We further assess the effectiveness of contextual residual
representations through comparisons with holistic feature
representations and ablations of masking strategies and
residual components.
Additional analyses investigate different editing scenarios,
robustness to common perturbations, and computational efficiency.

\noindent\textbf{Datasets and Evaluation Protocol.}\hspace{2mm}For AIGC video detection, we follow the {AIGVDBench}
protocol~\cite{ma2026your}, training on $13{,}785$ real videos
and $14{,}000$ videos generated by Open-Sora~\cite{zheng2024open}.
For AI-edited region localization, we use the OPN and STTN
subsets of {AINPAINT}~\cite{montibeller2026ainpaint},
each containing $1{,}146$ clips, yielding $2{,}292$ 16-frame
training clips.
The single-task variants are trained separately using their
respective datasets and task heads, whereas the unified VICA
is jointly trained on both datasets with a shared readout
and task-specific supervision.

For detection evaluation, we test on 20 open-source and
11 closed-source generators from AIGVDBench, alongside
{GenVidBench-143K}~\cite{ni2026genvidbench},
{VideoFeedback}~\cite{he2024videoscore},
{Magic Videos}~\cite{li2026preserving}, and
{GVF}~\cite{ma2025detecting} to assess cross-dataset
generalization.
For localization, we evaluate on eight AINPAINT editing
methods, including six unseen during training.
We assess cross-dataset generalization on MOSE100~\cite{lou2025video}
videos inpainted using E2FGVI, FuseFormer, and STTN.
For both benchmarks, full-length videos are processed using
non-overlapping 16-frame windows.

\noindent\textbf{Metrics.}\hspace{2mm}We evaluate AIGC video detection using video-level AUROC and AP (\%),
computed for each generator against the corresponding real videos
and averaged across generators.
For AI-edited region localization, we report token-level IoU and F1 (\%).
Ground-truth tokens are labeled as edited if their corresponding
tubelets contain at least one edited pixel, and predicted probabilities
are thresholded at $0.5$.
IoU and F1 are computed across all test videos for each editing method
and then averaged across methods.

\noindent\textbf{Baselines.}\hspace{2mm}We compare VICA against three categories of baselines.
For AIGC video detection, we include D3~\cite{zheng2025d3},
DeMamba~\cite{chen2026demamba}, DeCoF~\cite{ma2025detecting},
ReStraV~\cite{interno2026ai}, SPLIT~\cite{hyun2026split},
WaveRep~\cite{corvi2025seeing}, STALL~\cite{ben2026training},
and V-PVP~\cite{cui2026rethinking}.
We also consider AIGC image detectors, including
UniFD~\cite{ojha2023towards}, NPR~\cite{tan2024rethinking},
DDA~\cite{chen2025dual}, and Effort~\cite{yan2024effort},
which classify individual frames and aggregate predictions
by majority vote.
For AI-edited region localization, we compare against
ViLocal~\cite{lou2025video}, TruVIL~\cite{lou2025trusted},
AINPAINT~\cite{montibeller2026ainpaint}, and
RelayFormer~\cite{huang2026relayformer}.

\noindent\textbf{Implementation Details.}\hspace{2mm}
We employ a frozen V-JEPA~2.1 ViT-g/16 backbone and process
16-frame video clips at a resolution of $384\times384$.
Each clip is tokenized into $2\times16\times16$ tubelets,
yielding $T=8$ temporal steps with $N=24\times24$ spatial
tokens per step.
We extract $D=1408$-dimensional features from layers
$\{10,20,30,40\}$ and apply the complementary masking
strategy defined in \cref{eq:mask-ratio}.
The trainable components consist of a shared readout
($7.0$M parameters) and two lightweight task-specific heads,
with $7.01$M trainable parameters in total.

\noindent\textbf{Optimization and Inference.}\hspace{2mm}
We train VICA for $2{,}870$ steps (equivalent to $10$ epochs
on AINPAINT) using AdamW with an initial learning rate of
$10^{-3}$, a weight decay of $0.01$, a $5\%$ linear warmup,
and cosine learning rate decay.
Each training step includes $8$ AINPAINT clips and $12$
AIGVDBench videos ($6$ real and $6$ generated).
Data augmentations, including flipping, cropping, blurring,
and JPEG compression, are applied only to AINPAINT,
while no augmentation is applied to AIGVDBench.
At inference, videos are center-cropped and resized to
$384\times384$.
Detection uses 16-frame clips, while localization on both
AINPAINT and MOSE100 processes full-length videos using
non-overlapping 16-frame windows.
Further details on the shared readout architecture, data augmentation, and preprocessing are provided in Appendix~\ref{sec:suppl-readout}, while computational cost analysis is presented in Appendix~\ref{sec:suppl-cost}.

\subsection{Main Results}

\noindent\textbf{AIGC Video Detection.}\hspace{2mm}As shown in \cref{tab:core-main}, existing detectors exhibit considerable performance variations across different benchmarks, highlighting the challenges of generalizing forensic cues to unseen video generators.
DeCoF performs strongly on AIGVDBench's open-source generators but deteriorates substantially on GenVidBench and Magic Videos, while WaveRep struggles markedly on VideoFeedback.
Even V-PVP, which preserves fine-grained temporal information through patch-level aggregation, exhibits inconsistent performance across datasets.
These variations suggest that existing methods may capture generator-dependent forensic patterns whose discriminative power diminishes under distribution shifts.

By characterizing regional discrepancies through context-conditioned residuals, VICA explicitly incorporates broader spatiotemporal context into forensic analysis.
As shown in \cref{tab:core-main}, the single-task VICA-Det achieves the highest AUROC (94.74\%) on AIGVDBench's open-source generators, while substantially outperforming V-PVP on VideoFeedback and Magic Videos.
These improvements suggest that contextual residuals provide forensic cues beyond conventional temporal representations.
Joint training further enhances detection performance, with unified VICA achieving the highest AUROC on four of the six evaluation settings and the highest AP on five.
Notably, its AUROC on GVF increases from 64.55\% for VICA-Det to 87.01\%, highlighting the benefits of region-level localization supervision.
These results suggest that complementary supervision helps the shared residual readout learn more transferable forensic representations.
Additional robustness evaluations under common video perturbations are provided in Appendix~\ref{sec:suppl-robustness}, and detailed per-generator results are presented in Appendix~\ref{sec:suppl-generator}.

\begin{table}[t]
\centering
\caption{
All localization baselines and VICA-Loc are trained on the OPN
and STTN subsets of AINPAINT, while VICA-Unified additionally
uses AIGVDBench detection supervision.
Evaluation is conducted on the AINPAINT test set and MOSE100,
with full-length videos processed using non-overlapping
16-frame windows.
Token-level IoU and F1 are computed per editing method and
averaged across methods, while token-level false-positive
rate (FPR) is evaluated on authentic AINPAINT videos.
\textbf{Bold} and \underline{underlined} denote the best
and second-best results, respectively.
}
\vspace{-0.5em}
\label{tab:localization}
\setlength{\tabcolsep}{4.5pt}
\fontsize{6.5}{7.0}\selectfont
\setlength{\extrarowheight}{1.5pt}
\renewcommand{\arraystretch}{1.12}
\setlength{\aboverulesep}{0pt}
\setlength{\belowrulesep}{0pt}
\resizebox{\columnwidth}{!}{%
\begin{tabular}{l ccc cc}
\toprule
\multirow{2}{*}{Method}& \multicolumn{3}{c}{AINPAINT} & \multicolumn{2}{c}{MOSE100}\\
\cmidrule(lr){2-4}\cmidrule(lr){5-6}
 & IoU $\uparrow$ & F1 $\uparrow$ & FPR (\%) $\downarrow$ & IoU & F1\\
\midrule
ViLocal~\cite{lou2025video} & 12.31 & 20.93 & 4.97 & 18.75 & 30.85\\
TruVIL~\cite{lou2025trusted} & 14.80 & 24.38 & 6.94 & 33.84 & 50.42\\
AINPAINT~\cite{montibeller2026ainpaint} & 68.70 & 79.24 & \underline{1.07} & 64.14 & 78.15\\
RelayFormer~\cite{huang2026relayformer} & 57.16 & 69.23 & 1.32 & \underline{72.70} & \underline{81.70}\\
\midrule
\rowcolor{vicalight}VICA-Loc & \underline{70.76} & \underline{79.26} & 1.15 & 69.75 & 79.81\\
\rowcolor{vicaplus}VICA-Unified & \textbf{72.74} & \textbf{83.34} & \textbf{0.39} & \textbf{73.82} & \textbf{84.78}\\
\bottomrule
\end{tabular}%
}
\vspace{-2em}
\end{table}

\noindent\textbf{AI-Edited Region Localization.}\hspace{2mm}As shown in \cref{tab:localization}, existing localization methods exhibit substantial performance differences across editing datasets.
ViLocal and TruVIL perform relatively poorly on AINPAINT, while the specialized AINPAINT baseline achieves stronger results but shows reduced performance on MOSE100.
RelayFormer benefits from global--local contextual modeling and performs competitively on MOSE100, yet exhibits weaker performance on AINPAINT.
These variations highlight the challenges of transferring localized forensic cues across editing methods, suggesting that local artifact extraction or contextual information exchange alone may be insufficient for robust localization under distribution shifts.

By assessing regional contextual agreement, VICA provides transferable forensic evidence while preserving spatial specificity.
As shown in \cref{tab:localization}, the single-task VICA-Loc surpasses the specialized AINPAINT baseline on both AINPAINT and MOSE100 in IoU, supporting the effectiveness of context-conditioned residuals across editing settings.
Unified VICA achieves the highest IoU and F1 scores on both benchmarks, reaching 72.74\% IoU on AINPAINT and 73.82\% on MOSE100.
These improvements suggest that joint supervision benefits localization beyond task-specific training.
Moreover, unified VICA achieves a false-positive rate of 0.39\% on authentic AINPAINT videos, compared with 1.15\% for VICA-Loc, indicating improved discrimination between authentic and manipulated regions.
Together, these findings suggest that video-level supervision helps distinguish manipulation-related contextual discrepancies from natural prediction errors, improving localization reliability across editing datasets.

\subsection{Ablation Studies}

\noindent\textbf{Holistic vs. Residual Features.}\hspace{2mm}To assess the effectiveness of contextual agreement residuals, we compare VICA with holistic feature representations extracted from several frozen visual backbones.
All methods use the same training data, objectives, and task heads, with holistic baselines operating on features extracted from unmasked videos and VICA operating on prediction residuals from V-JEPA~2.1.
As shown in \cref{tab:backbone}, conventional visual backbones exhibit varying effectiveness across the two tasks, with holistic V-JEPA~2.1 achieving relatively strong localization performance (64.77\% IoU and 78.01\% F1) but lower detection performance (83.66\% AUROC and 84.13\% AP).
Replacing its holistic features with contextual agreement residuals improves detection by 9.86 and 9.36 percentage points in AUROC and AP, respectively, while increasing localization IoU and F1 by 7.97 and 5.33 points.
These consistent gains under the same pretrained backbone and training supervision demonstrate the effectiveness of residual-based representations for both tasks.
By retaining token-level prediction discrepancies, VICA preserves spatially localized evidence while incorporating information inferred from the broader spatiotemporal context.

\begin{table}[t]
\centering
\caption{Backbone comparison. All variants use frozen backbones
with identical task heads, training objectives, and data.
V-JEPA~2.1-based models extract features from four selected
layers, whereas other backbones use only the final layer.
Baselines operate on features extracted from unmasked videos,
while VICA uses context-conditioned prediction residuals
from V-JEPA~2.1.
We report video-level AUROC and AP for detection on AIGVDBench
and token-level IoU and F1 for localization on AINPAINT,
averaged across generators and editing methods, respectively.}
\label{tab:backbone}
\setlength{\tabcolsep}{4.5pt}
\setlength{\extrarowheight}{1.5pt}
\renewcommand{\arraystretch}{1.12}
\setlength{\aboverulesep}{0pt}
\setlength{\belowrulesep}{0pt}
\fontsize{6.5}{7.0}\selectfont
\resizebox{\columnwidth}{!}{%
\begin{tabular}{lcccc}
\toprule
\multirow{2}{*}{Backbone} & \multicolumn{2}{c}{Detection} & \multicolumn{2}{c}{Localization}\\
\cmidrule(lr){2-3}\cmidrule(lr){4-5}
 & AUROC & AP & IoU & F1\\
\midrule
CLIP ViT-L/14  & 86.77 & 86.81 & 39.06 & 50.24\\
% CLIP ViT-L/14 & Holistic & 86.77 & 86.81 & 39.06 & 50.24\\
SigLIP2 &  59.92 & 62.52 & 47.28 & 58.40\\
% DINOv2 ViT-L/14 & Holistic & 85.65 & 84.73 & 57.07 & 64.15\\
DINOv2 ViT-L/14 & 85.65 & 84.73 & 57.07 & 64.15\\
DINOv3 ViT-L & 85.10 & 84.73 & 59.15 & 65.14\\
VideoMAE & 75.66 & 74.52 & 16.69 & 23.67\\
VideoMAEv2 & 87.61 & 87.05 & 28.14 & 43.30\\
V-JEPA~2.1 & 83.66 & 84.13 & 64.77 & 78.01\\
\midrule
\rowcolor{vicaplus}VICA & 93.52 & 93.49 & 72.74 & 83.34 \\
\bottomrule
\end{tabular}%
}
\vspace{-2em}
\end{table}

\noindent\textbf{Mask Strategies.}\hspace{2mm}We investigate the impact of masking strategies on contextual residual representations in \cref{tab:mask}.
We compare uniform random masking at ratios of 10\%, 30\%, and 50\% with three non-uniform strategies.
Whole-segment masking hides the two temporal halves in complementary views, while the stepped schedule $(0,0,50,50,50,50,100,100)$ progressively increases masking ratios over time.
VICA adopts the two-level schedule $(20,20,20,20,80,80,80,80)$ defined in \cref{eq:mask-ratio}.
All strategies employ complementary masks to ensure complete regional coverage.

As shown in \cref{tab:mask}, VICA's two-level schedule achieves the best performance across both tasks, highlighting the importance of spatiotemporal context distribution.
Uniform random masking does not explicitly encourage temporally distant predictions, while whole-segment masking removes all visible information from one temporal half, potentially limiting prediction quality.
The proposed schedule balances these extremes by preserving visible spatial information at every temporal step while encouraging the use of temporally distant context.
These results suggest that effective contextual residuals depend not only on regional coverage but also on the distribution of visible tokens.

% \begin{table}[t]
% \centering
% \caption{Mask strategy ablation. Ratios give the percentage of masked tokens in each of the eight temporal maps for the first view. For non-uniform strategies, the second view masks the complementary tokens. Detection reports AUROC and AP (\%) on AIGVDBench, averaged over unseen generators. Localization reports token-level IoU and F1 (\%) on AINPAINT, averaged over editors. Results are averaged over three seeds.}
% \label{tab:mask}
% \setlength{\tabcolsep}{4pt}
% \resizebox{\columnwidth}{!}{%
% \begin{tabular}{llcccc}
% \toprule
%  & & \multicolumn{2}{c}{Detection} & \multicolumn{2}{c}{Localization}\\
% \cmidrule(lr){3-4}\cmidrule(lr){5-6}
% Strategy & Per-map mask ratio (\%) & AUROC & AP & IoU & F1\\
% \midrule
% \multicolumn{6}{l}{\textit{Uniform random}}\\
% Random 10\% & $(10,10,10,10,10,10,10,10)$ & 92.12 & 92.08 & 72.48 & 83.24 \\
% Random 30\% & $(30,30,30,30,30,30,30,30)$ & 92.60 & 92.18 & 72.73 & 83.13\\
% Random 50\% & $(50,50,50,50,50,50,50,50)$ & 92.68 & 92.59 & 72.09 & 82.62\\
% \midrule
% \multicolumn{6}{l}{\textit{Non-uniform}}\\
% Whole-segment & $(100,100,100,100,0,0,0,0)$ & 92.55 & 92.73 & 73.11 & 83.53\\
% Stepped & $(0,0,50,50,50,50,100,100)$ & 93.52 & 93.39 & 72.77 & 83.25\\
% Step (VICA) & $(20,20,20,20,80,80,80,80)$ & 93.52 & 93.49 & 73.82 & 84.78\\
% \bottomrule
% \end{tabular}%
% }
% \end{table}

\begin{table}[t]
\centering
\caption{Mask strategy ablation. We compare different masking strategies
on AIGVDBench and AINPAINT while keeping all other components fixed.
For detection, we report video-level AUROC and AP (\%) on AIGVDBench,
computed per open-source generator and averaged across generators.
For localization, we report token-level IoU and F1 (\%) on AINPAINT,
computed per editing method and averaged across methods.}
\vspace{-0.5em}
\label{tab:mask}
\setlength{\tabcolsep}{4.5pt}
\fontsize{6.5}{7.0}\selectfont
\resizebox{\columnwidth}{!}{%
\begin{tabular}{lcccc}
\toprule
\multirow{2}{*}{Strategy} & \multicolumn{2}{c}{Detection} & \multicolumn{2}{c}{Localization}\\
\cmidrule(lr){2-3}\cmidrule(lr){4-5}
 & AUROC & AP & IoU & F1\\
\midrule
\multicolumn{5}{l}{\textit{Uniform random}}\\
% Random 10\% & 92.12 & 92.08 & 72.48 & 83.24\\
% Random 30\% & 92.60 & 92.18 & 72.73 & 83.13\\
% Random 50\% & 92.68 & 92.59 & 72.09 & 82.62\\
% \midrule
% \multicolumn{5}{l}{\textit{Non-uniform}}\\
% Whole-segment & 92.55 & 92.73 & 73.11 & 83.53\\
% Stepped & 93.52 & 93.39 & 72.77 & 83.25\\
% Step (VICA) & 93.52 & 93.49 & 72.74 & 83.34\\
Random 10\% & 88.12 & 88.08 & 68.48 & 79.24\\
Random 30\% & 88.60 & 88.18 & 68.73 & 80.13\\
Random 50\% & 90.68 & 90.59 & 70.09 & 80.62\\
\midrule
\multicolumn{5}{l}{\textit{Non-uniform}}\\
Whole-segment & 86.55 & 86.73 & 66.11 & 76.53\\
Stepped & 88.52 & 89.39 & 69.77 & 79.25\\
Step (VICA) & 93.52 & 93.49 & 72.74 & 83.34\\
\bottomrule
\end{tabular}%
}
\vspace{-1.6em}

\end{table}

\begin{table}[t]
\centering
\caption{Residual ablation. We compare different combinations of spatial
and bidirectional temporal residuals while keeping the masking
strategy, task heads, and training data fixed.
$S$, $F$, and $B$ denote the spatial, forward, and backward
residuals, respectively, with the full VICA shown in the last row.
We report video-level AUROC (\%) for detection on AIGVDBench
and VideoFeedback, and token-level IoU and F1 (\%) for
localization on AINPAINT.}
%\vspace{-0.3em}

\label{tab:residual}
\setlength{\tabcolsep}{4.5pt}
\resizebox{\columnwidth}{!}{%
\begin{tabular}{ccc cc cc}
\toprule
\multicolumn{3}{c}{Residual} & \multicolumn{2}{c}{Detection} & \multicolumn{2}{c}{Localization}\\
\cmidrule(lr){1-3}\cmidrule(lr){4-5}\cmidrule(lr){6-7}
$S$ & $F$ & $B$ & AIGVDBench & VideoFeedback & IoU & F1\\
\midrule
\checkmark & & & 88.77 & 86.76 & 68.31 & 78.17\\
% \checkmark & & & 95.77 & 94.76 & 71.31 & 82.17\\
 & \checkmark & & 89.87 & 87.09 & 70.25 & 79.97\\
 & & \checkmark & 87.89 & 86.00 & 71.51 & 78.51\\
 & \checkmark & \checkmark & 90.27 & 87.47 & 70.63 & 79.12\\
\checkmark & \checkmark & & 90.87 & 88.75 & 70.53 & 80.07\\
\checkmark & & \checkmark & 90.07 & 88.30 & 70.70 & 80.93\\
\midrule
\checkmark & \checkmark & \checkmark & 93.52 & 90.69 & 72.74 & 83.34\\
\bottomrule
\end{tabular}%
}
%\vspace{-1.3em}

\end{table}
\begin{figure}[t]
    \centering
    \includegraphics[width=1.0\columnwidth]{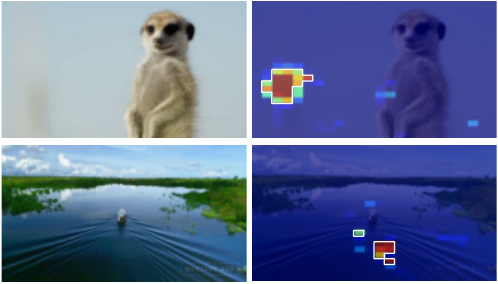}\
    %\vspace{-1.2em}
    \caption{Localization results on authentic videos. Each row presents an input frame (left) and its corresponding
per-region edit-probability overlay (right).
Sparse false-positive responses occur primarily in
featureless background regions rather than on foreground subjects.}
    \label{fig:false-positive}
    \vspace{-1.5em}
\end{figure}

% 时间，空间残差
\noindent\textbf{Residual Comparison.}\hspace{2mm}We investigate the contributions of spatial and bidirectional
temporal residuals in \cref{tab:residual}, keeping the masking
strategy, readout architecture, and training data fixed.
The spatial residual $S$ characterizes same-step prediction
discrepancies, while the forward and backward residuals
$F$ and $B$ incorporate temporal variations through
cross-step prediction--observation comparisons.
Individually, these residuals provide useful but incomplete
forensic cues, with their effectiveness varying across
detection and localization.
Despite the increased feature dimensionality, combining all three residuals, $S{+}F{+}B$, achieves the best performance on both tasks, suggesting the complementary value of spatial and bidirectional temporal forensic cues.

\noindent\textbf{False Positives on Real Videos.}\hspace{2mm}
We also examine VICA's localization responses on authentic
videos to understand false positives.
As illustrated in \cref{fig:false-positive}, the localization
head occasionally assigns non-negligible edit probabilities
to unmodified regions.
These false activations are sparse and primarily occur
in featureless background areas, such as the sky beside
the meerkat and the water near the boat, rather than
on foreground subjects.
Such regions may provide limited visual information for
reliable contextual prediction, leading to residual discrepancies
even in authentic content.
These observations reinforce that contextual disagreement
does not necessarily indicate manipulation, as natural
prediction errors can produce residual responses.
The learned localization head must therefore distinguish
manipulation-related discrepancies from natural prediction errors.
% \begin{figure}[t]
%     \centering
%     \fbox{\rule{0pt}{0.45\columnwidth}\rule{0.9\linewidth}{0pt}}
%     \caption{\textbf{False activations on real videos.} Localization responses of VICA on unedited clips. Some regions with large natural prediction error, such as fast motion or occlusion boundaries, receive non-trivial edit probabilities, although the responses are weaker and more scattered than on edited regions. [Placeholder, figure to be added.]}
%     \label{fig:false-positive}
% \end{figure}

% \subsection{Robustness and Computational Cost}
% Following the perturbation categories of TRACE~\cite{cao2026trace}, evaluate JPEG quality $\{100,90,80,70,60\}$, Gaussian blur $\sigma\in\{0,0.5,1,1.5,2\}$, and resize factors $\{0.5,1,1.5,2\}$ followed by restoration to the original size.
% Apply perturbations to decoded frames before each method's preprocessing and do not retrain on the test corruptions.
% Report detection AUROC, localization IoU, and performance on authentic clips with high motion or occlusion.
% Tabulate latency per 16-frame clip, peak memory, trainable parameters, and throughput on the same hardware, including all five encoder routes and four predictor passes for VICA.
% Batched parallel execution may reduce latency but does not remove the additional computation relative to Global-only.
\section{Conclusion}
\label{sec:conclusion}
We introduced VICA, a unified framework for AI-generated video detection and AI-edited region localization based on contextual agreement. By comparing regional observations with predictions inferred from the surrounding spatiotemporal context, VICA constructs spatial and bidirectional temporal residuals that characterize contextual discrepancies. These residuals provide a shared forensic representation for both tasks, shifting the focus from isolated manipulation artifacts to regional consistency with broader contextual expectations. Extensive experiments across multiple benchmarks demonstrate strong detection and localization performance, with robust generalization to unseen video generators and editing methods. Comparisons with features from the same pretrained backbone further validate the effectiveness of context-conditioned residuals, while joint supervision enhances their utility across both tasks. We discuss limitations and future work in Appendix~\ref{sec:suppl-limitation}.

% \section{Limitations}
% \label{sec:limitations}
% VICA has several limitations.
% First, the single-task variants are less robust than the unified model. VICA-Det, trained on detection data alone, degrades on some benchmarks such as GVF, where it trails detectors that model frame-to-frame dynamics. Joint training recovers this gap, as the unified VICA attains the best GVF result in \cref{tab:core-main}, but this indicates that the residual representation benefits from, and partly relies on, supervision from both tasks.
% Second, VICA depends on a frozen V-JEPA~2.1 model, so its evidence is bounded by what this backbone predicts, and errors in the predicted features propagate into the residuals.
% Third, the model operates on fixed 16-frame clips with fixed preprocessing, which may weaken cues that span longer durations or that are sensitive to resolution.
% Finally, computing the observed and predicted features requires multiple encoder and predictor passes, adding computation relative to a single holistic forward pass.
% Adapting the backbone, handling variable-length inputs, and reducing the prediction cost are left for future work.

\section*{LLM Usage Statement}

We used ChatGPT to assist with language polishing and proofreading. All scientific content, experimental results, and conclusions were developed and verified by the authors, who take full responsibility for the final manuscript.

{
    \small
    \bibliographystyle{ieeenat_fullname}
    \bibliography{main}
}

\maketitlesupplementary

\renewcommand{\thesection}{\Alph{section}}
\setcounter{section}{0}

\makeatletter
\setlength{\@dblfptop}{0pt}
\setlength{\@dblfpsep}{14pt plus 2pt minus 2pt}
\setlength{\@dblfpbot}{0pt plus 1fil}
\makeatother
\setcounter{dbltopnumber}{3}
\renewcommand{\dbltopfraction}{0.85}
\renewcommand{\textfraction}{0.12}

\section{Shared Readout \& Data Processing}
\label{sec:suppl-readout}

\begin{figure*}[!tp]
\centering
\includegraphics[width=\textwidth]{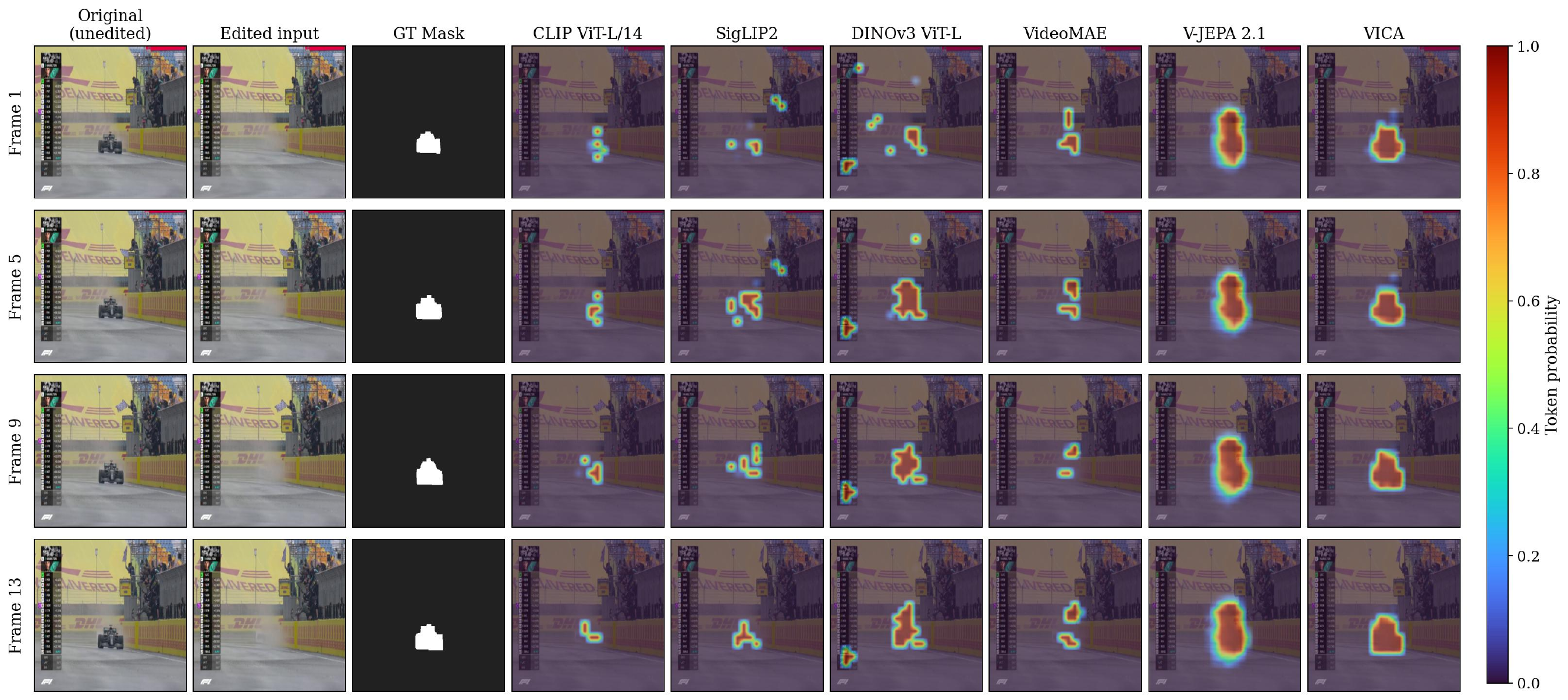}
\caption{An AI editor removed a Formula 1 car from this clip; the white masks mark its original location in each frame. With the same readout architecture, the five backbones using holistic features miss parts of the edited region or highlight the background. VICA's predictions more closely follow the ground-truth masks across all four frames, supporting the use of contextual residuals for localization.}
\label{fig:holistic_vica}
\end{figure*}

\noindent\textbf{Shared Readout Architecture}\hspace{2mm}
We use a lightweight convolutional--axial transformer readout $g$, shared by both tasks, while keeping the V-JEPA 2.1 backbone frozen. 
For a 16-frame $384 \times 384$ clip, each of the four feature levels $l = 1, \dots, 4$ yields $8 \times 24 \times 24$ spatial residuals $S^{l}$ ($D$-dimensional) and $4 \times 24 \times 24$ forward and backward temporal residuals $F^{l}, B^{l}$, one pair per two consecutive steps. 
Both steps of a pair share its $F^{l}, B^{l}$, so every step holds $[S^{l}, F^{l}, B^{l}] \in \mathbb{R}^{3D}$ per position. Concatenating the four levels gives the residual grid $R \in \mathbb{R}^{T \times H \times W \times 12D}$ with $T \times H \times W = 8 \times 24 \times 24$. 
The readout first fuses these residuals at the feature level: each slice $R^{l} \in \mathbb{R}^{T \times H \times W \times 3D}$ is projected to $C = 256$ channels by its own stem ($\text{LayerNorm} \to \text{Linear} \to \text{GELU}$), and a linear layer combines the four outputs into $H^{(0)} \in \mathbb{R}^{T \times H \times W \times C}$, one compact feature per token. 
To let residuals interact across space and time, three (2+1)D context blocks ($i = 1, 2, 3$) then refine $H^{(0)}$. Each block applies a spatial convolution with dilation $d_i \in \{1, 2, 4\}$, a temporal convolution, and a residual MLP, so the receptive field of every token grows with depth:
\begin{equation}
\begin{aligned}
    \tilde{H}^{(i)} &= \mathrm{DWConv1D}_{t}\!\left(\mathrm{DWConv2D}_{s,d_i}\!\left(H^{(i-1)}\right)\right), \\
    H^{(i)} &= H^{(i-1)} + \mathrm{MLP}\!\left(\mathrm{LN}\!\left(\tilde{H}^{(i)}\right)\right).
\end{aligned}
\end{equation}
Here $\mathrm{DWConv2D}_{s,d_i}$ is a depthwise $3 \times 3$ convolution over $(H, W)$ with dilation $d_i$, $\mathrm{DWConv1D}_{t}$ is a depthwise convolution over $T$ (kernel size $3$), $\mathrm{LN}$ is LayerNorm, and $\mathrm{MLP}$ is a two-layer GELU MLP ($C \to 2C \to C$). 
Two axial attention blocks then add global context, each applying spatial self-attention among the $H \times W$ tokens of each step, temporal self-attention among the $T$ steps at each position (8 heads each), and an MLP, all with pre-LayerNorm and residual connections. 
A final LayerNorm gives $Z = g(R)$, whose token $z_{t,n} \in \mathbb{R}^{C}$ is the feature of tubelet $(t, n)$ (Tab.~\ref{tab:readout_arch}). 
The localization head applies a linear classifier to each $z_{t,n}$ to predict whether the tubelet is manipulated. 
The detection head takes the mean and standard deviation of $z_{t,n}$ over the $H \times W$ positions of each step ($2C = 512$ dimensions), pools over the $T$ steps with log-sum-exp, and applies a linear classifier to predict the clip-level logit. 
The readout and both heads have $7.01\text{M}$ trainable parameters in total. Fig.~\ref{fig:holistic_vica} further compares localization using VICA's contextual residuals with localization using holistic features from five backbones, all with the same readout architecture.

\noindent\textbf{Data Processing}\hspace{2mm}
For detection, we uniformly sample $16$ of the first $\min(n,128)$ native-rate frames, center-crop to a square, and resize bicubically to $384\times384$ without augmentation.
For localization, we resize AINPAINT frames ($432\times240$) and masks directly to $384\times384$ using bilinear and nearest-neighbor interpolation, respectively. At test time, we process full AINPAINT and MOSE100 videos in non-overlapping $16$-frame windows.
During localization training, frames and masks share random horizontal flips and crops ($p=0.5$ each, crop area $60\%$--$100\%$). Only frames receive Gaussian blur ($p=0.3$, radius $0.1$--$1.5$) and JPEG compression ($p=0.5$, quality $40$--$100$).

\section{Computational Cost}
\label{sec:suppl-cost}

All experiments are conducted on $4$ NVIDIA L20 ($46\text{ GB}$) GPUs using PyTorch and CUDA under \texttt{bfloat16} precision. Only the lightweight readout and task heads are updated ($7.01\text{M}$ parameters), completing $10$ epochs in $1.8$ hours across the $4$ GPUs. During inference on the same $4$ GPUs, VICA processes each clip in $402\text{ ms}$ on average, reaching an aggregate throughput of $10.0\text{ clips/s}$ with a peak memory usage of $9.2\text{ GB}$ per GPU.

\section{Robustness}
\label{sec:suppl-robustness}

\begin{figure*}[!tp]
\centering
\includegraphics[width=0.99\textwidth]{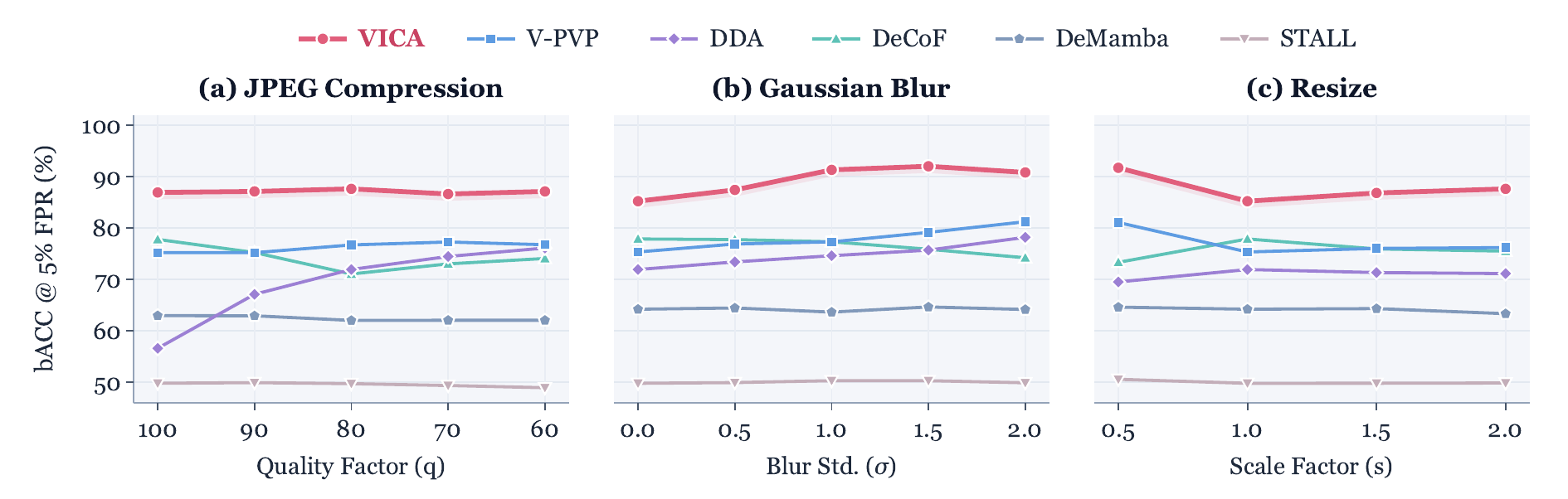}
\caption{Balanced accuracy at 5\% FPR on AIGVDBench under (a) JPEG compression, (b) Gaussian blur, and (c) resolution scaling.}
\label{fig:robustness_vica}
\end{figure*}

\noindent\textbf{Frame Quality Degradation.}\hspace{2mm}
We evaluate the robustness of VICA on AIGVDBench under three common image quality distortions: JPEG compression (quality $q \in \{100, 90, 80, 70, 60\}$), Gaussian blur (standard deviation $\sigma \in \{0.0, 0.5, 1.0, 1.5, 2.0\}$), and resolution scaling (scale $s \in \{0.5, 1.0, 1.5, 2.0\}$, restored to the original size). 
We distort raw frames before preprocessing and report balanced accuracy at a $5\%$ false-positive rate (bACC@5\%FPR) for all 14 settings.
We compare VICA with four AI-generated video detectors (V-PVP, DeCoF, DeMamba, and STALL) and with DDA, the best-performing image detector on AIGVDBench in our main experiments. VICA outperforms all five baselines across the 14 settings in Fig.~\ref{fig:robustness_vica}. Its bACC stays near $86\%$ across JPEG quality levels and peaks near $91\%$ at $\sigma=1.5$ under Gaussian blur. Among the four resize settings, downsampling to half resolution ($s=0.5$) yields the highest bACC, about $91\%$.

% % (optional, Video encoder) TODO://Video Encoding H264, CRF

\begin{figure*}[!tp]
\centering
\includegraphics[width=0.99\textwidth]{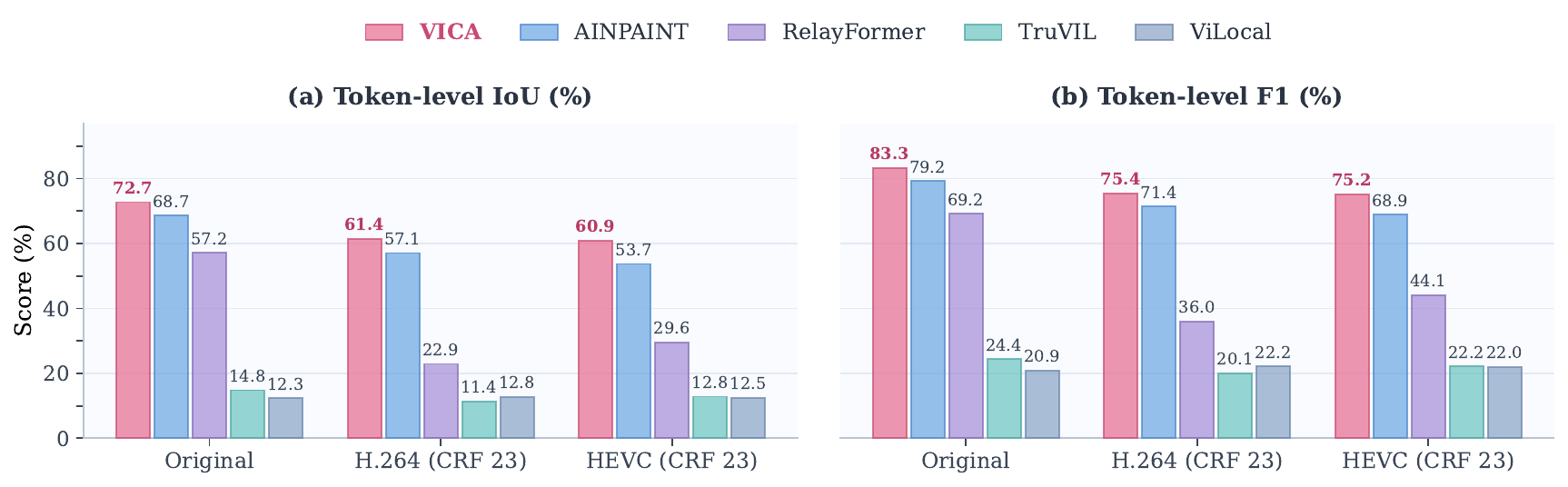}
\caption{Token-level (a) IoU and (b) F1 on AINPAINT test videos: original, H.264, and HEVC (both at CRF~23).}
\label{fig:video_compression}
\end{figure*}

\noindent\textbf{Video Compression.}\hspace{2mm}
We follow the compression setting of AINPAINT and re-encode the AINPAINT test videos with H.264 and HEVC at CRF~23 (Fig.~\ref{fig:video_compression}).
Compression lowers the accuracy of the leading methods. VICA's IoU falls from $72.7\%$ to $61.4\%$ with H.264 and to $60.9\%$ with HEVC, and its F1 falls from $83.3\%$ to $75.4\%$ and $75.2\%$.
VICA still has the best IoU and F1 under both codecs. Compared with the second-best method, AINPAINT, it is higher by $4.3$ IoU and $4.0$ F1 points with H.264, and by $7.2$ IoU and $6.3$ F1 points with HEVC.
In terms of the drop after compression, VICA and AINPAINT lose a comparable amount (about $11$ to $15$ IoU points), while RelayFormer loses $28$ to $34$ IoU points, suggesting that it is much more sensitive to compression.

% % (optional, Video encoder) TODO://Video Encoding H264, CRF

\section{Detailed Generator Performance}
\label{sec:suppl-generator}

% P0, TODO:// 定性分析 observed feature和Predicted Feature在不同Mask策略下 到底有什么区别，从distance-level/similarity-level

\noindent\textbf{Detection.}\hspace{2mm}
We evaluate detection on AIGVDBench's 20 open-source generators, comparing each generator's output with the corresponding real videos using AUROC (Tab.~\ref{tab:per_gen_open_auroc}) and bACC@5\%FPR (Tab.~\ref{tab:per_gen_open_bacc}). Unified VICA averages $93.52\%$ AUROC and $83.48\%$ bACC. Its AUROC exceeds $90\%$ for all image-to-video (I2V) generators and 7 of 12 text-to-video (T2V) generators, and reaches $88.98\%$ and $85.76\%$ for the two video-to-video (V2V) generators. VICA-Det scores higher than unified VICA on both metrics for the T2V outputs of CogVideoX 1.5, IPOC, and LTX. It also ranks first on average, with $94.74\%$ AUROC and $85.25\%$ bACC.

\noindent\textbf{Localization.}\hspace{2mm}
Across the eight editors in the AINPAINT test set (Tab.~\ref{tab:per_editor_ainpaint}), VICA achieves the highest mean token-level IoU ($72.74\%$) and F1 ($83.34\%$), exceeding VICA-Loc by $1.98$ and $4.08$ points, respectively. VICA leads on both metrics for CoCoCo, GMCNN, and OPN. Relative to VICA-Loc, its IoU gains $15.87$ points on CoCoCo but drops $11.53$ points on LDVI.

\section{Limitations}
\label{sec:suppl-limitation}
Our current design has two limitations. First, VICA-Det outperforms the unified VICA on some generators, suggesting that mixing localization and detection data does not uniformly improve detection across all generators. The relative amount of each type of supervision may therefore require careful tuning to find a better balance. Second, VICA relies on the quality of V-JEPA 2.1 predictions and requires both observed and predicted features. The target encoder processes the unmasked clip to obtain observed features. Predicting features for all tokens also requires encoding and running the predictor on two complementary masked views, which adds computation relative to using observed features alone. We will investigate the supervision balance and computational efficiency in future work.

\begin{table*}[!tp]
\centering
\caption{Architecture of the shared readout and task-specific heads.}
% Output shapes are shown for a $16$-frame $384 \times 384$ clip ($T = 8$, $H = W = 24$). All attention output projections in the axial blocks are zero-initialized.
\label{tab:readout_arch}
\small
\setlength{\tabcolsep}{2.2pt}
\setlength{\extrarowheight}{1.5pt}
\renewcommand{\arraystretch}{1.12}
\setlength{\aboverulesep}{0pt}
\setlength{\belowrulesep}{0pt}
\begin{tabular}{llcc}
\toprule
Stage & Layer Specification & Output Shape & Params \\
\midrule
Per-Level Stem ($\times 4$) & $\text{LN}(4224) \to \text{Linear}(4224, 256) \to \text{GELU}$ & $4 \times (8, 24, 24, 256)$ & 4.36M \\
Level Fusion & $\text{Concat} \to \text{Linear}(1024, 256)$ & $(8, 24, 24, 256)$ & 0.26M \\
Context Blocks ($\times 3$) & $\text{DWConv2D}_{3\times 3,\, d\in\{1,2,4\}} \to \text{DWConv1D}_{3} \to \text{LN} \to \text{MLP}$ & $(8, 24, 24, 256)$ & 0.80M \\
Axial Attn. ($\times 2$) & $\text{Spatial Attn (8h)} \to \text{Temporal Attn (8h)} \to \text{MLP}$ & $(8, 24, 24, 256)$ & 1.58M \\
Output Norm & $\text{LayerNorm}(256)$ & $(8, 24, 24, 256)$ & $<$0.01M \\
\midrule
Localization Head & $\text{Linear}(256, 1)$ & $(8, 24, 24, 1)$ & $<$0.01M \\
Detection Head & $\text{Spatial [Mean; Std]} \to \text{Temp. LSE Pool} \to \text{Linear}(512, 1)$ & $(1,)$ & $<$0.01M \\
\midrule
\textbf{Total Trainable} & Shared Readout ($7.00\text{M}$) + Task Heads & -- & \textbf{7.01M} \\
\bottomrule
\end{tabular}
\end{table*}

\begin{table*}[!tp]
\centering
\caption{\textbf{Open-Source} per-generator AUROC (\%) results on the 20 open-source video generators of AIGVDBench (3,000 real vs.\ 3,000 generated videos per generator). \textbf{Bold} and \underline{underlined} denote the best and second-best results, respectively.}
\label{tab:per_gen_open_auroc}

\def\gh#1{{\scriptsize\renewcommand{\arraystretch}{0.92}\begin{tabular}{@{}c@{}}\rule{0pt}{2.6ex}#1\rule[-1.0ex]{0pt}{0pt}\end{tabular}}}
\setlength{\tabcolsep}{2pt}
\setlength{\extrarowheight}{1.5pt}
\renewcommand{\arraystretch}{1.12}
\setlength{\aboverulesep}{0pt}
\setlength{\belowrulesep}{0pt}
\resizebox{\textwidth}{!}{%
\begin{tabular}{c|*{6}{>{\centering\arraybackslash}m{1.08cm}}|*{12}{>{\centering\arraybackslash}m{1.08cm}}|*{2}{>{\centering\arraybackslash}m{1.08cm}}|>{\centering\arraybackslash}m{1.08cm}}
\toprule
\multicolumn{22}{c}{\textbf{Open-Source Approaches}} \\
\midrule
\multirow{2.5}{*}{\textbf{Method}} &
\multicolumn{6}{c|}{\textbf{I2V}} &
\multicolumn{12}{c|}{\textbf{T2V}} &
\multicolumn{2}{c|}{\textbf{V2V}} &
\multirow{2.5}{*}{\textbf{AVG}} \\
\cmidrule(lr){2-7} \cmidrule(lr){8-19} \cmidrule(lr){20-21}
& \gh{Easy\\Animate} & \gh{LTX} & \gh{Pyramid\\Flow} & \gh{SEINE} & \gh{SVD} & \gh{Video\\Crafter}
& \gh{Acc\\Video} & \gh{Animate\\Diff} & \gh{Cogvideo\\x1.5} & \gh{Easy\\Animate} & \gh{Hunyuan} & \gh{IPOC} & \gh{LTX} & \gh{Open\\Sora} & \gh{Pyramid\\Flow} & \gh{Rep\\Video} & \gh{Video\\Crafter} & \gh{Wan\\2.1}
& \gh{Cogvideo\\x1.5} & \gh{LTX} & \\
\midrule
\multicolumn{22}{c}{\textbf{AI-Generated Image Detection Models}} \\
\midrule
NPR          & 56.93 & 49.48 & 49.76 & 42.40 & 52.24 & 51.73 & 65.08 & 48.27 & 54.56 & 67.53 & 64.52 & 60.36 & 59.62 & 45.42 & 74.64 & 62.62 & 53.10 & 77.08 & 49.11 & 50.83 & 56.76 \\
DDA          & \underline{93.43} & 83.98 & 89.15 & 82.51 & 87.07 & \underline{99.94} & 78.36 & 82.75 & 78.76 & \underline{91.44} & 86.65 & 94.28 & 87.82 & 88.37 & 86.86 & \textbf{94.95} & \underline{99.97} & \textbf{99.98} & \textbf{93.87} & 89.18 & 89.47 \\
UniFD        & 52.89 & 60.26 & 53.20 & 75.57 & 72.93 & 75.48 & 49.06 & 50.99 & 53.06 & 49.04 & 47.20 & 54.06 & 64.02 & 48.60 & 54.01 & 55.80 & 57.07 & 50.66 & 53.04 & 53.04 & 56.50 \\
Effort       & 60.66 & 72.43 & 83.82 & 92.51 & 96.12 & 50.93 & 51.22 & 55.33 & 78.38 & 56.47 & 53.40 & 86.88 & 82.19 & \underline{99.98} & 95.76 & 71.32 & 63.67 & 54.11 & 70.66 & 64.82 & 72.03 \\
\midrule
\multicolumn{22}{c}{\textbf{AI-Generated Video Detection Models}} \\
\midrule
D3           & 55.16 & 55.97 & 60.16 & 56.32 & 50.55 & 72.22 & 74.33 & 82.47 & 63.14 & 67.06 & 74.37 & 65.00 & 55.74 & 64.77 & 63.72 & 61.25 & 83.18 & 60.82 & 68.73 & 62.74 & 64.88 \\
DeMamba      & 62.60 & 66.87 & 78.62 & 92.32 & 96.65 & 50.12 & 65.11 & 61.37 & 75.49 & 69.49 & 65.52 & 82.13 & 79.72 & \textbf{100.00} & 88.77 & 74.68 & 64.98 & 52.49 & 59.35 & 62.33 & 72.43 \\
DeCoF        & 83.92 & 85.93 & \underline{91.92} & 98.61 & 98.67 & 86.13 & 84.12 & 87.15 & \underline{92.81} & 89.84 & 82.21 & \underline{96.12} & \underline{96.08} & \textbf{100.00} & \textbf{98.91} & \underline{92.52} & 96.66 & 71.39 & 85.63 & 83.43 & 90.10 \\
ReStraV      & \textbf{97.37} & 52.63 & 52.14 & 39.63 & 49.19 & 27.61 & 39.14 & 15.21 & 70.47 & \textbf{92.52} & 45.37 & 84.69 & 48.96 & 96.87 & 47.51 & 91.06 & 17.41 & 71.95 & 49.57 & 51.17 & 57.02 \\
SPLIT        & 36.16 & 45.23 & 29.34 & 88.60 & 76.11 & 79.59 & 27.60 & 63.76 & 50.85 & 25.34 & 28.62 & 15.96 & 30.47 & 39.76 & 25.11 & 21.22 & 44.07 & 46.09 & 19.29 & 33.42 & 41.33 \\
WaveRep      & 62.28 & 75.09 & 89.35 & 94.11 & 98.77 & 55.31 & 68.91 & 56.16 & 80.59 & 68.89 & 72.15 & 79.23 & 89.42 & \textbf{100.00} & 93.35 & 75.45 & 69.37 & 64.49 & 66.34 & 68.31 & 76.38 \\
STALL        & 49.87 & 53.17 & 52.13 & 54.97 & 48.72 & 44.77 & 42.36 & 31.76 & 43.48 & 37.19 & 38.80 & 37.19 & 53.72 & 39.21 & 44.46 & 44.30 & 34.59 & 41.08 & 49.37 & 50.78 & 44.60 \\
V-PVP        & 72.95 & 89.17 & 89.37 & 92.37 & 94.99 & 90.84 & \textbf{91.88} & 93.80 & 83.58 & 79.13 & \textbf{89.93} & 86.42 & 88.25 & \textbf{100.00} & 94.33 & 85.32 & 98.11 & 70.61 & 85.85 & \underline{89.98} & 88.34 \\
\midrule
\rowcolor{vicalight}
VICA-Det     & 91.14 & \textbf{96.68} & \textbf{93.82} & \textbf{99.98} & \textbf{99.94} & \textbf{100.00} & \underline{90.63} & \textbf{99.81} & \textbf{95.69} & 87.03 & \underline{87.75} & \textbf{97.20} & \textbf{96.69} & \textbf{100.00} & \underline{98.21} & 89.38 & \textbf{100.00} & 89.04 & \underline{93.31} & \textbf{93.77} & \textbf{94.74} \\
\rowcolor{vicaplus}
VICA-Unified & 93.20 & \underline{93.65} & 91.74 & \underline{99.86} & \underline{99.59} & \textbf{100.00} & 89.91 & \underline{99.24} & 89.15 & 90.00 & 83.02 & 93.52 & 94.03 & \textbf{100.00} & 96.63 & 89.42 & \textbf{100.00} & \underline{92.78} & 88.98 & 85.76 & \underline{93.52} \\
\bottomrule
\end{tabular}%
}
\end{table*}

\begin{table*}[!tp]
\centering
\caption{\textbf{Open-Source} per-generator bACC@5\%FPR (\%) results on the 20 open-source video generators of AIGVDBench (3,000 real vs.\ 3,000 generated videos per generator). \textbf{Bold} and \underline{underlined} denote the best and second-best results, respectively.}
\label{tab:per_gen_open_bacc}
\setlength{\tabcolsep}{2.2pt}
\setlength{\extrarowheight}{1.5pt}
\renewcommand{\arraystretch}{1.12}
\setlength{\aboverulesep}{0pt}
\setlength{\belowrulesep}{0pt}
\def\gh#1{{\scriptsize\renewcommand{\arraystretch}{0.92}\begin{tabular}{@{}c@{}}\rule{0pt}{2.6ex}#1\rule[-1.0ex]{0pt}{0pt}\end{tabular}}}
\resizebox{\textwidth}{!}{%
\begin{tabular}{c|*{6}{>{\centering\arraybackslash}m{1.08cm}}|*{12}{>{\centering\arraybackslash}m{1.08cm}}|*{2}{>{\centering\arraybackslash}m{1.08cm}}|>{\centering\arraybackslash}m{1.08cm}}
\toprule
\multicolumn{22}{c}{\textbf{Open-Source Approaches}} \\
\midrule
\multirow{2.5}{*}{\textbf{Method}} &
\multicolumn{6}{c|}{\textbf{I2V}} &
\multicolumn{12}{c|}{\textbf{T2V}} &
\multicolumn{2}{c|}{\textbf{V2V}} &
\multirow{2.5}{*}{\textbf{AVG}} \\
\cmidrule(lr){2-7} \cmidrule(lr){8-19} \cmidrule(lr){20-21}
& \gh{Easy\\Animate} & \gh{LTX} & \gh{Pyramid\\Flow} & \gh{SEINE} & \gh{SVD} & \gh{Video\\Crafter}
& \gh{Acc\\Video} & \gh{Animate\\Diff} & \gh{Cogvideo\\x1.5} & \gh{Easy\\Animate} & \gh{Hunyuan} & \gh{IPOC} & \gh{LTX} & \gh{Open\\Sora} & \gh{Pyramid\\Flow} & \gh{Rep\\Video} & \gh{Video\\Crafter} & \gh{Wan\\2.1}
& \gh{Cogvideo\\x1.5} & \gh{LTX} & \\
\midrule
\multicolumn{22}{c}{\textbf{AI-Generated Image Detection Models}} \\
\midrule
NPR          & 52.68 & 50.35 & 50.18 & 48.12 & 50.37 & 52.58 & 62.60 & 50.23 & 53.67 & 61.20 & 60.55 & 57.10 & 55.92 & 48.65 & 64.08 & 57.50 & 52.67 & 67.58 & 50.55 & 50.95 & 54.88 \\
DDA          & 77.50 & 57.93 & 54.37 & 66.73 & 79.32 & \textbf{99.83} & 61.35 & 69.92 & 59.30 & 74.55 & 64.02 & 80.07 & 62.73 & 58.78 & 54.22 & \underline{82.57} & \textbf{99.93} & \textbf{99.97} & \underline{75.33} & 60.42 & 71.94 \\
UniFD        & 51.90 & 58.97 & 53.07 & 70.58 & 67.17 & 71.72 & 50.58 & 51.55 & 53.12 & 50.08 & 49.15 & 53.87 & 61.78 & 49.80 & 53.80 & 54.77 & 56.30 & 50.87 & 53.25 & 52.98 & 55.76 \\
Effort       & 60.55 & 72.03 & \textbf{83.15} & 91.40 & 94.78 & 50.90 & 51.18 & 55.22 & 77.80 & 56.38 & 53.33 & 85.97 & 81.47 & \textbf{98.28} & \underline{94.50} & 70.92 & 63.40 & 54.05 & 70.28 & 64.60 & 71.51 \\
\midrule
\multicolumn{22}{c}{\textbf{AI-Generated Video Detection Models}} \\
\midrule
D3           & 50.32 & 54.68 & 51.37 & 50.65 & 48.40 & 58.67 & 63.10 & 68.35 & 55.98 & 57.10 & 61.38 & 58.28 & 53.20 & 50.75 & 54.85 & 54.60 & 66.47 & 53.93 & 56.77 & 53.77 & 56.13 \\
DeMamba      & 55.47 & 59.42 & 66.65 & 83.88 & 90.80 & 50.50 & 54.15 & 54.17 & 64.78 & 57.20 & 55.78 & 70.63 & 67.22 & \underline{97.50} & 76.20 & 63.22 & 55.33 & 51.17 & 54.25 & 56.00 & 64.21 \\
DeCoF        & 66.02 & 70.00 & 77.70 & 93.80 & 94.77 & 69.20 & 66.27 & 72.88 & \underline{80.95} & 73.10 & 64.52 & \underline{87.58} & \underline{87.12} & \underline{97.50} & \textbf{95.00} & 79.98 & 88.75 & 56.85 & 69.62 & 66.30 & 77.89 \\
ReStraV      & \textbf{92.10} & 50.10 & 50.33 & 47.65 & 48.12 & 47.58 & 48.35 & 47.53 & 57.33 & \textbf{84.77} & 48.75 & 78.68 & 51.30 & 88.43 & 49.43 & \textbf{84.73} & 47.50 & 58.48 & 50.13 & 49.63 & 59.05 \\
SPLIT        & 51.17 & 51.88 & 48.17 & 66.90 & 59.42 & 59.68 & 50.45 & 54.43 & 52.28 & 49.28 & 49.38 & 48.12 & 49.95 & 48.88 & 48.43 & 48.50 & 49.27 & 51.72 & 47.92 & 48.43 & 51.71 \\
WaveRep      & 57.55 & 65.03 & 79.92 & 88.32 & 94.52 & 52.07 & 56.37 & 53.42 & 66.07 & 58.08 & 58.63 & 65.92 & 79.95 & \underline{97.50} & 83.75 & 62.18 & 56.75 & 54.88 & 58.13 & 58.72 & 67.39 \\
STALL        & 50.07 & 52.11 & 51.12 & 52.40 & 50.82 & 49.34 & 48.89 & 48.05 & 49.35 & 48.14 & 48.29 & 49.02 & 52.26 & 48.58 & 49.89 & 49.39 & 47.50 & 48.76 & 50.64 & 51.00 & 49.78 \\
V-PVP        & 58.87 & 72.85 & 75.97 & 81.98 & 87.65 & 76.82 & \textbf{77.48} & 80.55 & 68.90 & 63.93 & \textbf{72.65} & 72.97 & 73.22 & \underline{97.50} & 85.35 & 70.25 & 92.78 & 55.97 & 67.43 & \underline{74.17} & 75.36 \\
\midrule
\rowcolor{vicalight}
VICA-Det     & 77.50 & \textbf{87.70} & \underline{82.00} & \textbf{97.50} & \textbf{97.48} & \underline{97.50} & 72.60 & \textbf{97.22} & \textbf{86.17} & 70.03 & \underline{68.37} & \textbf{89.75} & \textbf{88.48} & \underline{97.50} & 92.85 & 75.28 & \underline{97.50} & 72.40 & \textbf{78.68} & \textbf{78.52} & \textbf{85.25} \\
\rowcolor{vicaplus}
VICA-Unified & \underline{81.36} & \underline{82.71} & 78.26 & \underline{97.24} & \underline{96.51} & \underline{97.50} & \underline{75.38} & \underline{95.56} & 75.46 & \underline{76.26} & 67.49 & 81.94 & 82.36 & \underline{97.50} & 89.07 & 75.86 & \underline{97.50} & \underline{80.38} & 72.70 & 68.61 & \underline{83.48} \\
\bottomrule
\end{tabular}%
}
\end{table*}

\begin{table*}[!tp]
\centering
\caption{\textbf{Closed-Source} per-generator AUROC (\%) results on the 11 closed-source video generators of AIGVDBench (2,000 real videos per generator). \textbf{Bold} and \underline{underlined} denote the best and second-best results, respectively.}
\label{tab:per_gen_closed_auroc}
\small
\setlength{\tabcolsep}{3pt}
\setlength{\extrarowheight}{1.5pt}
\renewcommand{\arraystretch}{1.12}
\setlength{\aboverulesep}{0pt}
\setlength{\belowrulesep}{0pt}
\def\gh#1{{\scriptsize\renewcommand{\arraystretch}{0.92}\begin{tabular}{@{}c@{}}\rule{0pt}{2.6ex}#1\rule[-1.0ex]{0pt}{0pt}\end{tabular}}}
\begin{tabular}{c|*{11}{>{\centering\arraybackslash}m{1.05cm}}|>{\centering\arraybackslash}m{1.05cm}}
\toprule
\multirow{2.5}{*}{\textbf{Method}} &
\multicolumn{11}{c|}{\textbf{Closed-Source Approaches}} &
\multirow{2.5}{*}{\textbf{AVG}} \\
\cmidrule(lr){2-12}
& \gh{Gen2} & \gh{Gen3} & \gh{Jimeng} & \gh{Luma} & \gh{Open\\Sora} & \gh{Sora} & \gh{Causvid\\24fps} & \gh{Kling} & \gh{Pika} & \gh{Vidu} & \gh{Wan} & \\
\midrule
\multicolumn{13}{c}{\textbf{AI-Generated Image Detection Models}} \\
\midrule
NPR    & 70.26 & 66.97 & 86.85 & 69.11 & 59.52 & 66.70 & 79.45 & 61.73 & 61.77 & 55.75 & 67.03 & 67.74 \\
DDA    & 93.02 & \underline{89.00} & \underline{92.46} & \underline{87.37} & 90.06 & \textbf{90.17} & \underline{91.38} & 82.44 & 89.87 & 65.52 & \textbf{90.33} & 87.42 \\
UniFD  & 49.33 & 56.79 & 61.77 & 57.34 & 48.76 & 49.76 & 56.96 & 52.47 & 57.30 & 51.69 & 49.52 & 53.79 \\
Effort & 52.38 & 61.91 & 54.55 & 56.87 & 79.70 & 61.12 & 78.27 & 63.49 & 57.73 & 83.25 & 61.33 & 64.60 \\
\midrule
\multicolumn{13}{c}{\textbf{AI-Generated Video Detection Models}} \\
\midrule
D3      & 88.63 & 50.22 & 68.40 & 69.21 & 75.63 & 67.48 & 69.12 & 68.60 & 81.55 & 72.13 & 75.86 & 71.53 \\
DeMamba & 60.95 & 66.07 & 55.93 & 51.61 & 67.78 & 61.22 & 70.42 & 63.66 & 63.60 & 66.35 & 63.87 & 62.86 \\
DeCoF   & 60.89 & 70.01 & 46.45 & 68.70 & 91.16 & 74.66 & \textbf{98.78} & 80.43 & 83.13 & 84.93 & 74.88 & 75.82 \\
ReStraV & 23.70 & 42.91 & 18.47 & 37.56 & 45.00 & 54.35 & 49.16 & 43.14 & 39.41 & 43.19 & 51.36 & 40.75 \\
SPLIT   & 22.55 & 62.06 & 34.30 & 41.07 & 32.79 & 60.32 & 56.11 & 46.08 & 32.77 & 46.81 & 45.18 & 43.64 \\
WaveRep & 85.18 & 88.15 & 79.88 & 82.84 & 88.07 & \underline{83.86} & 88.07 & 77.98 & 86.54 & 81.31 & 79.04 & 83.72 \\
STALL   & 79.68 & 75.19 & 77.98 & 79.44 & 81.68 & 76.59 & 79.42 & 72.75 & 79.75 & 72.04 & 74.79 & 77.21 \\
V-PVP   & \underline{97.96} & 77.10 & 92.24 & 85.39 & \underline{96.83} & 81.08 & 74.73 & \underline{87.75} & \underline{98.27} & \textbf{91.41} & \underline{88.32} & \underline{88.28} \\
\midrule
\rowcolor{vicalight}
VICA-Det & 96.23 & 83.63 & 83.20 & 78.66 & \textbf{97.80} & 77.58 & 73.28 & 85.37 & \textbf{98.79} & \underline{88.51} & 81.74 & 85.89 \\
\rowcolor{vicaplus}
VICA-Unified     & \textbf{99.78} & \textbf{89.92} & \textbf{95.00} & \textbf{89.70} & 95.66 & 81.66 & 84.49 & \textbf{91.27} & 96.73 & 81.57 & 85.67 & \textbf{90.13} \\
\bottomrule
\end{tabular}
\end{table*}

\begin{table*}[!tp]
\centering
\caption{\textbf{Closed-Source} per-generator bACC@5\%FPR (\%) results on the 11 closed-source video generators of AIGVDBench (2,000 real videos per generator). \textbf{Bold} and \underline{underlined} denote the best and second-best results, respectively.}
\label{tab:per_gen_closed_bacc}
\small
\setlength{\tabcolsep}{3pt}
\setlength{\extrarowheight}{1.5pt}
\renewcommand{\arraystretch}{1.12}
\setlength{\aboverulesep}{0pt}
\setlength{\belowrulesep}{0pt}
\def\gh#1{{\scriptsize\renewcommand{\arraystretch}{0.92}\begin{tabular}{@{}c@{}}\rule{0pt}{2.6ex}#1\rule[-1.0ex]{0pt}{0pt}\end{tabular}}}
\begin{tabular}{c|*{11}{>{\centering\arraybackslash}m{1.05cm}}|>{\centering\arraybackslash}m{1.05cm}}
\toprule
\multirow{2.5}{*}{\textbf{Method}} &
\multicolumn{11}{c|}{\textbf{Closed-Source Approaches}} &
\multirow{2.5}{*}{\textbf{AVG}} \\
\cmidrule(lr){2-12}
& \gh{Gen2} & \gh{Gen3} & \gh{Jimeng} & \gh{Luma} & \gh{Open\\Sora} & \gh{Sora} & \gh{Causvid\\24fps} & \gh{Kling} & \gh{Pika} & \gh{Vidu} & \gh{Wan} & \\
\midrule
\multicolumn{13}{c}{\textbf{AI-Generated Image Detection Models}} \\
\midrule
NPR     & 70.36 & 61.40 & 79.65 & 66.63 & 60.16 & 66.48 & \underline{81.63} & 63.24 & 64.08 & 56.43 & 67.73 & 67.07 \\
DDA     & \underline{81.49} & \textbf{77.67} & \underline{79.88} & \underline{74.39} & 77.64 & \textbf{80.47} & 72.95 & \underline{76.49} & 79.60 & 56.57 & \textbf{81.82} & \underline{76.27} \\
UniFD   & 50.62 & 53.35 & 59.42 & 54.64 & 49.49 & 50.09 & 54.17 & 51.97 & 56.89 & 51.34 & 50.64 & 52.97 \\
Effort  & 48.56 & 56.70 & 49.73 & 54.41 & 73.13 & 55.61 & 71.28 & 57.90 & 52.63 & \textbf{75.46} & 55.28 & 59.15 \\
\midrule
\multicolumn{13}{c}{\textbf{AI-Generated Video Detection Models}} \\
\midrule
D3      & 70.05 & 49.53 & 53.65 & 54.67 & 61.82 & 55.20 & 50.12 & 54.69 & 64.87 & 58.52 & 59.57 & 57.52 \\
DeMamba & 51.85 & 60.92 & 50.42 & 52.57 & 58.30 & 55.30 & 56.12 & 55.61 & 57.30 & 55.97 & 55.52 & 55.44 \\
DeCoF   & 49.97 & 54.63 & 47.67 & 55.27 & 70.35 & 54.83 & \textbf{84.97} & 61.36 & 61.98 & 61.05 & 55.57 & 59.79 \\
ReStraV & 47.85 & 49.00 & 47.95 & 47.95 & 48.40 & 48.82 & 47.67 & 47.88 & 48.37 & 48.25 & 48.45 & 48.24 \\
SPLIT   & 47.95 & 52.71 & 49.47 & 50.28 & 50.57 & 55.33 & 49.20 & 53.35 & 49.32 & 51.73 & 53.77 & 51.24 \\
WaveRep & 63.32 & 69.46 & 58.82 & 64.27 & 70.47 & \underline{67.22} & 66.00 & 63.40 & 66.85 & 65.17 & 64.95 & 65.45 \\
STALL   & 69.65 & 62.12 & 66.25 & 67.25 & 69.92 & 63.38 & 62.02 & 59.08 & 66.12 & 58.58 & 64.47 & 64.44 \\
V-PVP   & 81.02 & 55.17 & 68.50 & 61.12 & \underline{82.33} & 54.70 & 49.32 & 63.85 & \underline{87.30} & 65.12 & 62.05 & 66.41 \\
\midrule
\rowcolor{vicalight}
VICA-Det     & 72.97 & 58.07 & 55.88 & 53.12 & 81.88 & 51.77 & 48.72 & 65.05 & 80.40 & 60.98 & 54.77 & 62.15 \\
\rowcolor{vicaplus}
VICA-Unified & \textbf{97.52} & \underline{76.50} & \textbf{87.70} & \textbf{77.30} & \textbf{87.30} & 65.55 & 67.05 & \textbf{80.88} & \textbf{94.02} & \underline{65.32} & \underline{72.72} & \textbf{79.26} \\
\bottomrule
\end{tabular}
\end{table*}

% P0, TODO://补一个Loc 在AINPAINT的不同generator上的分别结果。Token-F1/IoU

\begin{table*}[!tp]
\centering
\caption{Per-editor token-level IoU / F1 (\%) results on the AINPAINT test set. \textbf{Bold} and \underline{underlined} denote the best and second-best results for each metric, respectively.}
\label{tab:per_editor_ainpaint}
\footnotesize
\setlength{\tabcolsep}{3pt}
\setlength{\extrarowheight}{1.5pt}
\renewcommand{\arraystretch}{1.12}
\setlength{\aboverulesep}{0pt}
\setlength{\belowrulesep}{0pt}
\resizebox{\textwidth}{!}{%
\begin{tabular}{c|cccccccc|c}
\toprule
\textbf{Method} & \textbf{CoCoCo} & \textbf{DSTT} & \textbf{FGVC} & \textbf{FuseFormer} & \textbf{GMCNN} & \textbf{LDVI} & \textbf{OPN} & \textbf{STTN} & \textbf{AVG} \\
\midrule
ViLocal               & 2.81\,/\,5.47 & 6.32\,/\,11.89 & 8.87\,/\,16.29 & 7.40\,/\,13.79 & 10.02\,/\,18.22 & 10.26\,/\,18.62 & 30.58\,/\,46.83 & 22.22\,/\,36.36 & 12.31\,/\,20.93 \\
TruVIL                & 4.67\,/\,8.92 & 21.13\,/\,34.89 & 6.76\,/\,12.67 & 12.84\,/\,22.75 & 6.91\,/\,12.93 & 8.75\,/\,16.09 & 39.80\,/\,56.94 & 17.55\,/\,29.85 & 14.80\,/\,24.38 \\
AINPAINT              & 24.20\,/\,38.97 & \textbf{83.78}\,/\,\textbf{91.18} & \textbf{69.25}\,/\,\textbf{81.83} & \textbf{82.70}\,/\,\textbf{90.53} & 81.14\,/\,\underline{89.59} & \underline{44.43}\,/\,\underline{61.53} & 81.86\,/\,90.03 & \textbf{82.24}\,/\,\textbf{90.26} & 68.70\,/\,79.24 \\
RelayFormer           & 18.06\,/\,30.60 & 79.92\,/\,88.84 & \underline{60.12}\,/\,75.09 & 76.11\,/\,86.43 & 32.45\,/\,48.99 & 29.33\,/\,45.35 & \underline{82.92}\,/\,\underline{90.66} & 78.37\,/\,87.87 & 57.16\,/\,69.23 \\
\midrule
\rowcolor{vicalight}
VICA-Loc     & \underline{48.11}\,/\,\underline{63.00} & \underline{82.35}\,/\,88.05 & 55.69\,/\,69.48 & 79.65\,/\,83.89 & \underline{84.47}\,/\,88.30 & \textbf{54.02}\,/\,\textbf{68.10} & 79.70\,/\,85.71 & 82.08\,/\,87.55 & \underline{70.76}\,/\,\underline{79.26} \\
\rowcolor{vicaplus}
VICA-Unified         & \textbf{63.98}\,/\,\textbf{78.13} & 81.62\,/\,\underline{89.66} & 59.64\,/\,\underline{75.27} & \underline{80.41}\,/\,\underline{89.07} & \textbf{86.24}\,/\,\textbf{92.38} & 42.49\,/\,60.25 & \textbf{85.43}\,/\,\textbf{92.35} & \underline{82.13}\,/\,\underline{89.62} & \textbf{72.74}\,/\,\textbf{83.34} \\
\bottomrule
\end{tabular}%
}
\end{table*}

\end{document}